\documentclass{article}

\usepackage{fullpage}
\usepackage{times}
\usepackage{amsfonts}
\usepackage{amsmath} 
\usepackage{amssymb}  

\usepackage{graphics} 
\usepackage{graphicx}
\usepackage{subfigure}

\usepackage{url}
\usepackage{enumitem}

\usepackage[linesnumbered,ruled,lined]{algorithm2e}

\usepackage[utf8]{inputenc}

\usepackage{comment}
\usepackage{multirow}			
\usepackage{hhline}
\usepackage{hyperref}

\newtheorem{example2}{\bf Example}
\newtheorem{definition}{Definition}
\newtheorem{execexample}{\bf Execution Example}

\newcommand{\rif}{\stackrel{\,\,+}{\leftarrow}}

\newcounter{ctr}

\title{Towards a Theory of Intentions for Human-Robot Collaboration}

\author{Rocio Gomez\\
  Electrical and Computer Engineering\\
  University of Auckland, NZ \\
  \texttt{rgom004@aucklanduni.ac.nz}
  \and
  Mohan Sridharan\\
  School of Computer Science\\
  University of Birmingham, UK\\
  \texttt{m.sridharan@bham.ac.uk}
  \and
  Heather Riley\\
  Electrical and Computer Engineering\\
  University of Auckland, NZ \\
  \texttt{hril230@aucklanduni.ac.nz}  
}

\date{}

\begin{document}

\maketitle

\begin{abstract}
  The architecture described in this paper encodes a theory of
  intentions based on the the key principles of non-procrastination,
  persistence, and automatically limiting reasoning to relevant
  knowledge and observations. The architecture reasons with transition
  diagrams of any given domain at two different resolutions, with the
  fine-resolution description defined as a refinement of, and hence
  tightly-coupled to, a coarse-resolution description.  Non-monotonic
  logical reasoning with the coarse-resolution description computes an
  activity (i.e., plan) comprising abstract actions for any given
  goal.  Each abstract action is implemented as a sequence of concrete
  actions by automatically zooming to and reasoning with the part of
  the fine-resolution transition diagram relevant to the current
  coarse-resolution transition and the goal.  Each concrete action in
  this sequence is executed using probabilistic models of the
  uncertainty in sensing and actuation, and the corresponding
  fine-resolution outcomes are used to infer coarse-resolution
  observations that are added to the coarse-resolution history.  The
  architecture's capabilities are evaluated in the context of a
  simulated robot assisting humans in an office domain, on a physical
  robot (Baxter) manipulating tabletop objects, and on a wheeled robot
  (Turtlebot) moving objects to particular places or people. The
  experimental results indicate improvements in reliability and
  computational efficiency compared with an architecture that does not
  include the theory of intentions, and an architecture that does not
  include zooming for fine-resolution reasoning.
\end{abstract}


\section{Introduction}
\begin{figure}[tb]
  \begin{center}
    \includegraphics[width=0.4\textwidth]{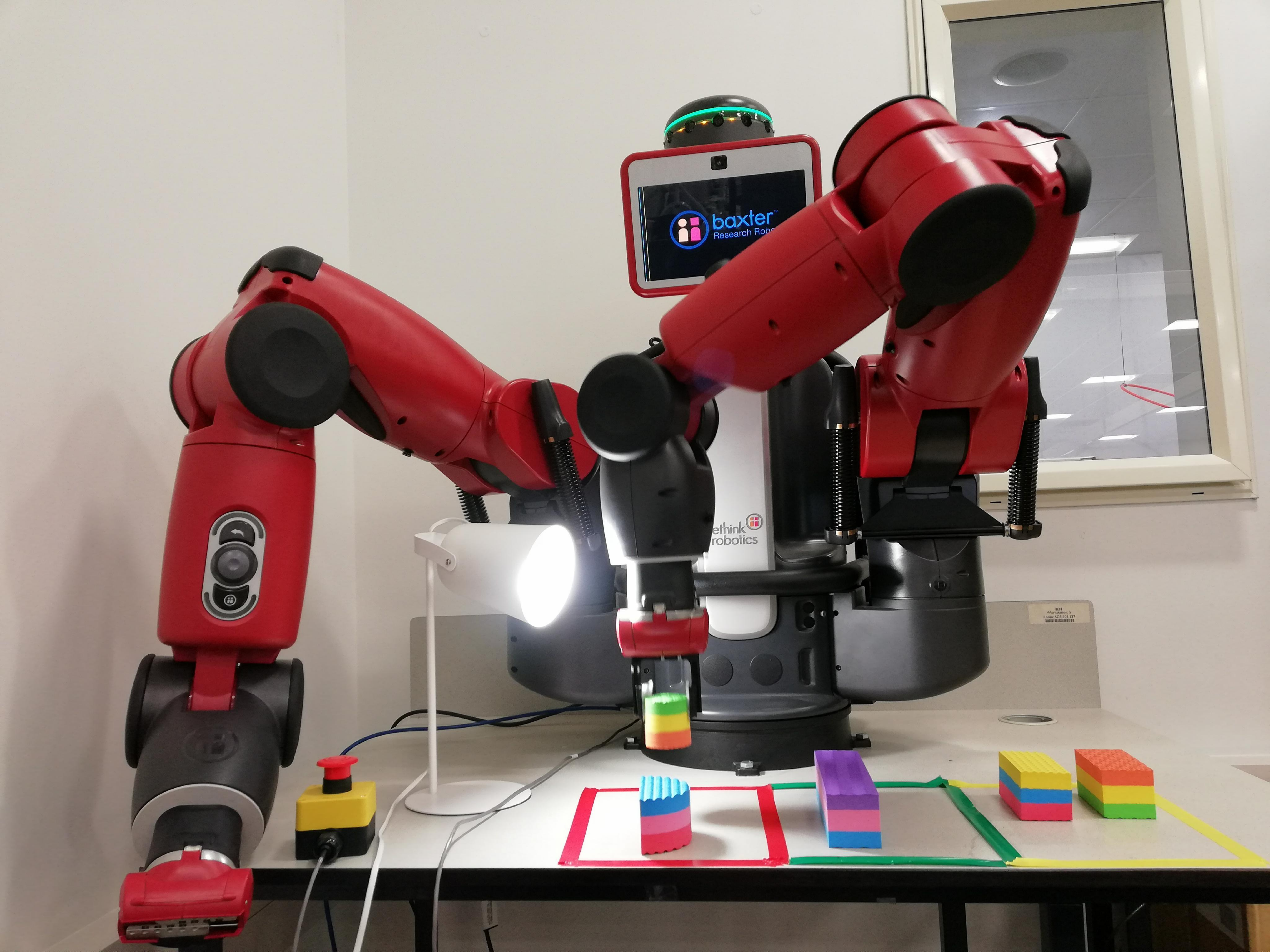}
    \hspace{1em}
    \includegraphics[width=0.3\textwidth]{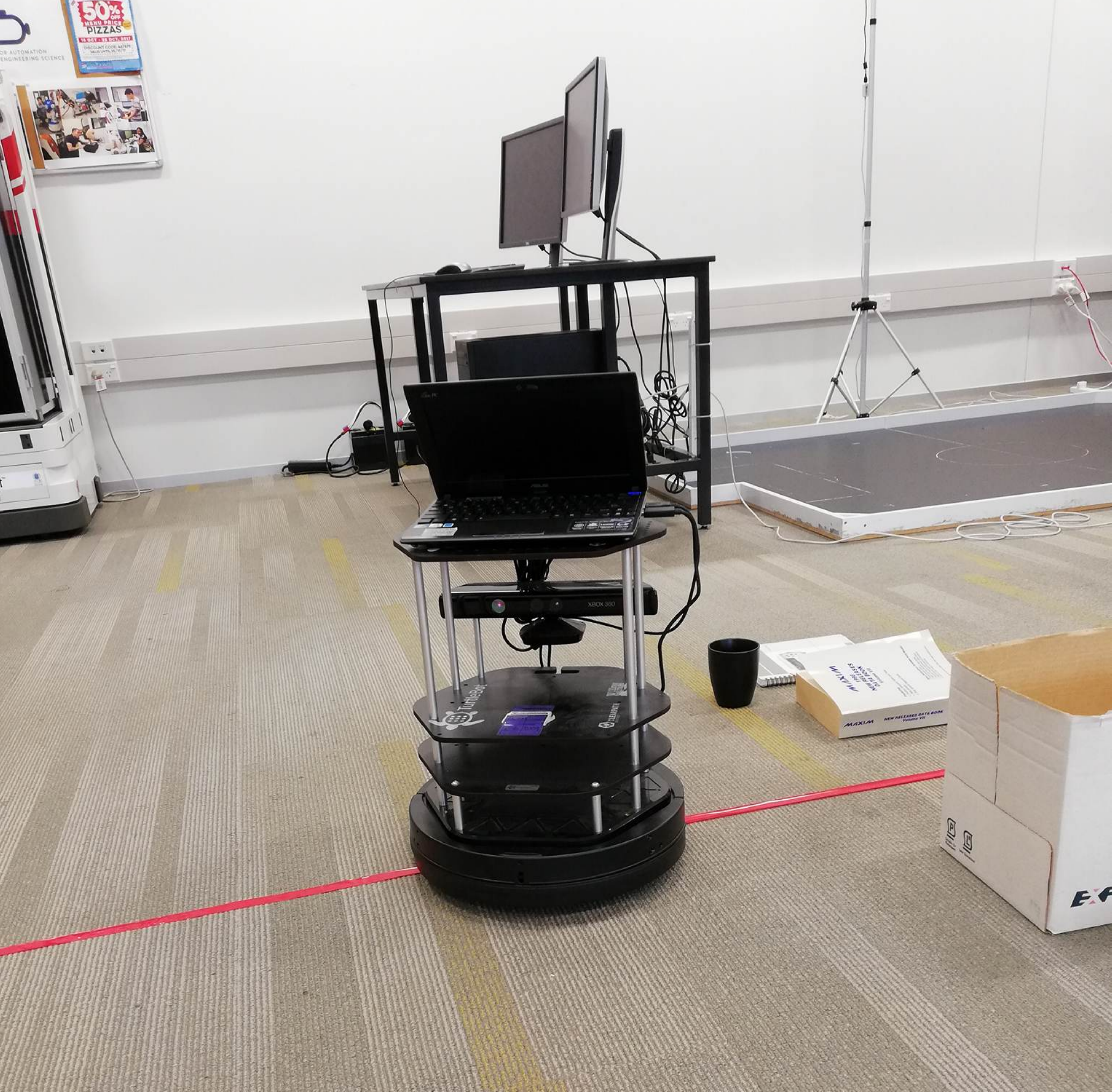}
  \end{center}
  \caption{Robot platforms used in the experimental trials reported in
    this paper: (a) Baxter robot manipulating objects on a tabletop;
    and (b) Turtlebot moving objects to particular locations in a
    lab.}
  \label{fig:example-robots}
\end{figure}

Consider a wheeled robot delivering objects to particular places or
people, or a robot with manipulators stacking objects in particular
configurations on a tabletop, as shown in
Figure~\ref{fig:example-robots}. Such robots that are deployed to
assist humans in dynamic domains have to reason with different
descriptions of uncertainty and incomplete domain knowledge.
Information about the domain often includes commonsense knowledge,
especially default knowledge that holds in all but a few exceptional
circumstances. For instance, the robot may be told that ``books are
usually in the library, but cookbooks may be in the kitchen''.  The
robot also extracts information from sensor inputs using algorithms
that quantify uncertainty probabilistically, e.g., ``I am $95\%$
certain the robotics book is on the table''. Although it is difficult
to equip robots with comprehensive domain knowledge or provide
elaborate supervision, reasoning with incomplete or incorrect
information can lead to incorrect or suboptimal outcomes, especially
when the robot is faced with unexpected success or failure. For
example, a robot may be asked to move two books from the office to the
library in a domain with four rooms. If this robot can only grasp one
object at a time, it will plan to move one book at a time from the
office to the library. After moving the first book, if the robot
observes the second book in the library, or in another room on the way
back to the office, it should stop executing the current plan because
this plan will no longer achieve the desired goal. Instead, it should
reason about this unexpected observation and compute a new plan if
necessary. One way to achieve this behavior with a traditional
planning system is to reason about all observations of domain objects
and events during plan execution, but this approach is computationally
unfeasible in complex domains.  The architecture described in this
paper, on the other hand, achieves the desired behavior by equipping a
robot pursuing a particular goal with an \emph{adapted theory of
  intentions}. This theory builds on the fundamental principles of
non-procrastination and persistence in the pursuit of a desired goal.
It enables the robot to reason about mental actions and states,
automatically identifying and considering the domain observations
relevant to the current action and the goal during planning and
execution. We refer to actions in such plans as \emph{intentional}
actions. We describe the following characteristics of our
architecture:
\begin{itemize}
\item The domain's transition diagrams at two different resolutions
  are described in an action language, with the fine-resolution
  transition diagram defined as a refinement of the coarse-resolution
  transition diagram. At the coarse resolution, non-monotonic logical
  reasoning with commonsense knowledge produces a sequence of
  intentional abstract actions for any given goal.

\item Each intentional abstract action is implemented as a sequence of
  concrete actions by automatically zooming to and reasoning with the
  part of the fine-resolution system description relevant to the
  current coarse-resolution transition and the goal. Each concrete
  action in this sequence is executed using probabilistic models of
  uncertainty, and the outcomes are added to the coarse-resolution
  history.
\end{itemize}
Action languages are formalisms that are used to model domain dynamics
(i.e., action effects). We chose to use an extension to action
language $\mathcal{AL}_d$~\cite{gelfond:ANCL13}, which we introduced
in prior work to model non-Boolean fluents and non-deterministic
causal laws~\cite{mohan:JAIR19}, because it provides the desired
expressive power for robotics domains. Also, we chose to translate our
action language descriptions to programs in
CR-Prolog~\cite{balduccini:aaaisymp03}, an extension of Answer Set
Prolog (ASP)~\cite{gelfond2014knowledge}, because it supports
non-monotonic logical reasoning with incomplete commonsense knowledge
in dynamic domains, which is a key desired capability in
robotics\footnote{We use the terms ``ASP'' and ``CR-Prolog''
  interchangeably in this paper. In the logic programming literature,
  ASP is also referred to as ``Answer Set Programming'' but we chose
  to use the earlier ``Answer Set Prolog'' expansion because we often
  refer to ASP programs.}. Furthermore, for the execution of each
concrete action, we use existing algorithms that include probabilistic
models of the uncertainty in perception and actuation.

Our architecture builds on the complementary strengths of prior work
on an architecture that used declarative programming to reason about
intended actions to achieve a given goal~\cite{blount2015theory}, and
an architecture that introduced step-wise refinement of
tightly-coupled transition diagrams at two different resolutions to
support non-monotonic logical reasoning and probabilistic reasoning
for planning and diagnostics~\cite{mohan:JAIR19}. Prior work on the
refinement-based architecture does not include a theory of intentions.
Also, prior work on the theory of intentions does not consider the
uncertainty in sensing and actuation, and does not scale to complex
domains. The key contributions of our architecture are thus to:
\begin{itemize}
\item enable planning with intentional abstract actions, and the
  associated mental states, actions, and beliefs, in the presence of
  incomplete domain knowledge, partial observability, and
  non-deterministic action outcomes; and
\item support scalability to larger domains by automatically
  restricting fine-resolution reasoning to knowledge and observations
  relevant to the goal or the coarse-resolution abstract action at
  hand, and by using probabilistic models of the uncertainty in
  sensing and actuation only when executing concrete actions.
\end{itemize}
We demonstrate the applicability of our architecture in the context of
a: (i) simulated robot assisting humans in an office domain; (ii)
physical robot (Baxter) manipulating objects on a tabletop; and (iii)
wheeled robot (Turtlebot) moving target objects to desired locations
or people in an office domain. We show that our architecture improves
reliability and computational efficiency in comparison with a baseline
architecture that does not reason about intentional actions and
beliefs at different resolutions, and with a baseline architecture
that does not limit reasoning to the relevant part of the domain.

The remainder of the paper is organized as follows. First,
Section~\ref{sec:relwork} reviews some related work to motivate the
need for our architecture. Section~\ref{sec:arch} then describes the
knowledge representation and reasoning architecture. The results of
evaluating the capabilities of this architecture are described in
Section~\ref{sec:expres}, followed by a description of the conclusions
and future work in Section~\ref{sec:conclusions}.

\section{Related Work}
\label{sec:relwork}
There is much work on modeling, recognizing, and reasoning about
intentions. For instance, Belief-desire-intention (BDI) architectures
model the intentions of reasoning agents and use these models to
eliminate choices that are inconsistent with the agent's current
intentions~\cite{bratman1987intention,rao:icmas95}. However, these
approaches do not learn from experience, are unable to adapt to new
situations, and make it difficult (by themselves) to explicitly
represent or reason about goals (e.g., for planning). There has been
work in developing probabilistic graphical models that enable a robot
to reason with encoded domain knowledge and learned models to
recognize a human participant's
intentions~\cite{kelley:TAMD12,kelley:hri08}. These approaches assume
that the structure of the models used to represent knowledge is known
a priori (e.g., the nodes and links of a hidden Markov model), and use
prior (observed) data to estimate the model parameters, e.g., the
probabilities of particular state transitions, and of obtaining
particular observations. Reasoning about intent, and identifying
discrepancies between expectations and observations, has also been
modeled as a component of architectures for agents that perform
goal-directed reasoning. For instance, a recent architecture models
metacognitive expectations by allowing agents to reason about their
cognition~\cite{dannenhauer:ACS18}. This meta-reasoning is achieved by
introducing different levels in the architecture, along with distinct
mechanisms at each level, to represent and reason about the domain
knowledge and the beliefs of the associated agent. Having such
separate levels that are not tightly coupled limits generalization,
and the smooth transfer of control and information between the levels.
 
Initial work on formalizing intentions based on declarative
programming introduced an action language and two fundamental
principles: (i) \emph{non-procrastination}, i.e., intended actions are
executed as soon as possible; and (ii) \emph{persistence}, i.e.,
unfulfilled intentions persist~\cite{baral2005reasoning}. This
architecture did not model agents with specific goals, but it was used
to enable an observer to recognize an agent's activity and
intention~\cite{gabaldon:ijcai09}. The \emph{Theory of Intentions}
($\mathcal{TI}$) extended this work to goal-driven agents by expanding
transition diagrams with physical states and physically executable
actions to include mental fluents and
actions~\cite{blount2014towards,blount2015theory}. It associated a
sequence of agent actions (called an ``activity'') with the goal it
intended to achieve, and the \emph{intentional agent} only performed
activities needed to achieve the goal. This theory has been used to
understand narratives of restaurant
scenarios~\cite{zhang2017application}, and to model goal-driven agents
in dynamic domains~\cite{saribatur2017reactive}. A requirement of such
theories is that the domain knowledge, including the preconditions and
effects of actions and goals, be encoded in advance, which is
difficult to do in robot domains. Also, the set of states (and
actions) can be large in robot domains, making efficient reasoning a
challenging task. Recent work attempted to improve computational
efficiency of reasoning with such theories by clustering
indistinguishable states~\cite{zeynep2016logics}, but this approach
required the clusters to be encoded in
advance~\cite{zhang2017application}.  Furthermore, these approaches do
not consider the uncertainty in sensing and actuation, which is the
primary source of error in robotics.

Logic-based methods have been used widely in robotics, including those
that also support probabilistic
reasoning~\cite{hanheide:AIJ17,zhang:TRO15}. Methods based on
classical first-order logic do not support non-monotonic logical
reasoning or the desired expressiveness, e.g., it is not always
meaningful to express degrees of belief by attaching probabilities to
logic statements. Logics such as ASP support non-monotonic logical
reasoning and have been used in cognitive
robotics~\cite{erdem2012applications} and many other
applications~\cite{erdem:AIM16}. However, classical ASP formulations
do not support probabilistic models of uncertainty, and such models
are used widely to model the uncertainty in sensing and actuation in
robotics. Approaches based on logic programming also do not support
one or more of the desired capabilities such as reasoning with large
probabilistic components, or incremental addition of probabilistic
information and variables to reason about open worlds.  As a step
towards addressing these challenges, our prior refinement-based
architecture reasoned with tightly-coupled transition diagrams at two
resolutions~\cite{mohan:JAIR19}. For any given goal, each abstract
action in a coarse-resolution plan computed using ASP-based reasoning
with commonsense knowledge, was executed as a sequence of concrete
actions computed by probabilistic reasoning over the relevant part of
the fine-resolution diagram using partially observable Markov decision
processes. In this paper, we explore the combination of the principles
of step-wise refinement with those of $\mathcal{TI}$. In comparison
with prior work, the architecture described in this paper supports
reasoning about intentional actions and beliefs in the presence of
incomplete domain knowledge, partial observability, and
non-deterministic action outcomes, and it incorporates a more
efficient approach for fine-resolution reasoning to support
scalability to larger domains.

\section{Knowledge Representation and Reasoning Architecture}
\label{sec:arch}

\begin{figure}[tb]
  \centering
  \includegraphics[width=0.8\textwidth]{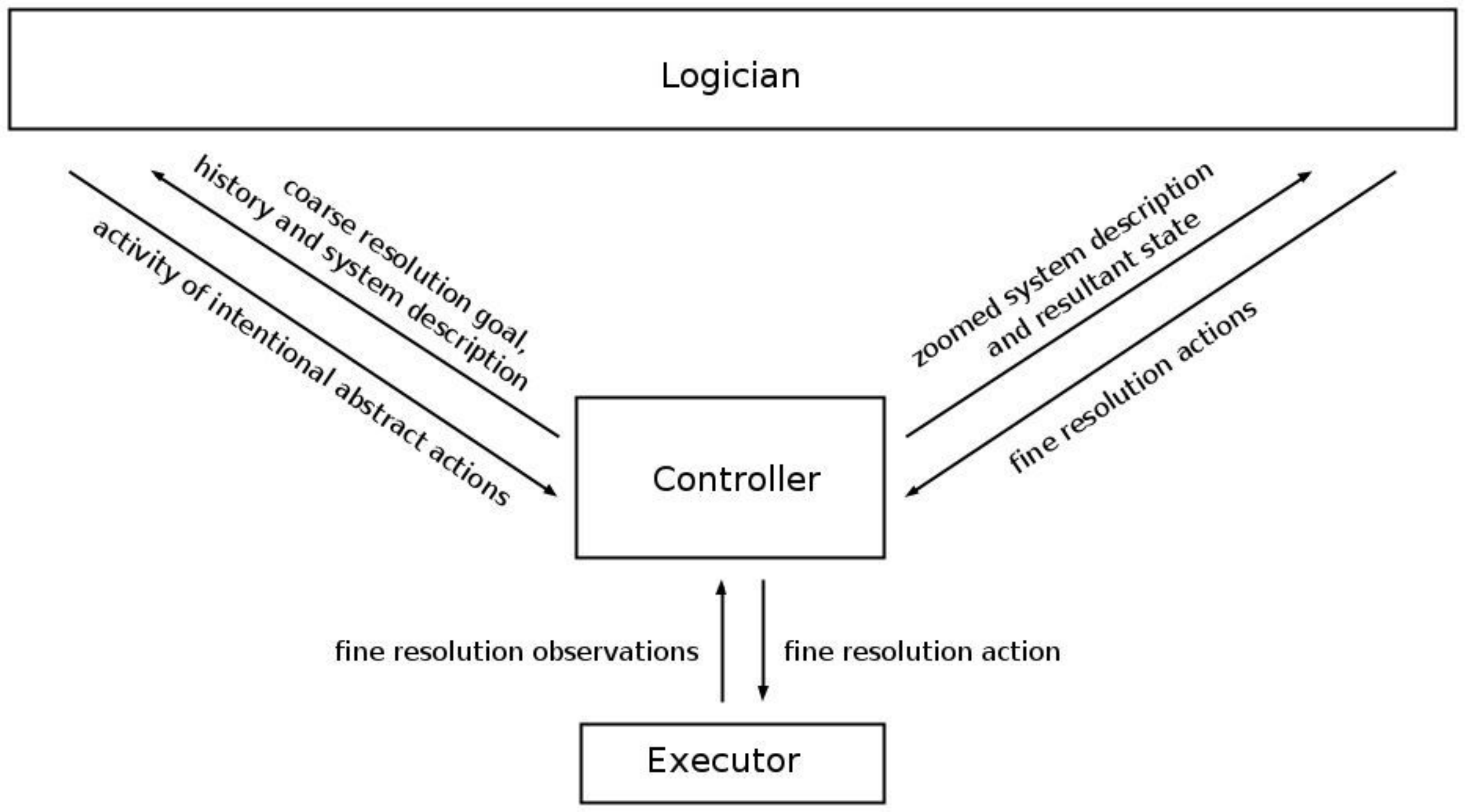}
  \caption{Architecture represents intentions and beliefs as tightly
    coupled transition diagrams at two different resolutions. It
    combines the complementary strengths of declarative programming
    and probabilistic reasoning, and may be viewed as interactions
    between a controller, logician, and executor.}
  \label{fig:arch-overview}
\end{figure}

Figure~\ref{fig:arch-overview} is a simplified block diagram of the
overall architecture. Similar to prior work~\cite{mohan:JAIR19}, this
architecture may be viewed as comprising three tightly-coupled
components: a controller, a logician, and an executor; the significant
differences in comparison with prior work are described later in this
section. The controller maintains the overall beliefs regarding the
state of the domain, and transfers control and information between the
components. Reasoning is based on transition diagrams of the domain at
two different resolutions, with a fine-resolution representation
defined as a refinement of a coarse-resolution representation of the
domain. For any given goal, the logician performs non-monotonic
logical reasoning with the coarse-resolution representation of
commonsense domain knowledge to generate an activity, i.e., a sequence
of intentional abstract actions to achieve the goal. To implement each
such intentional abstract action, the controller automatically zooms
to the part of the fine-resolution representation that is relevant to
the desired abstract transition and the goal. Reasoning with this
relevant part provides a plan of concrete actions; each such concrete
action is executed by the executor using probabilistic models of the
uncertainty in sensing and actuation. The observed and inferred
outcomes of executing a concrete action, along with any other relevant
observations, are communicated to the controller and added to the
coarse-resolution history. The logician reasons with this history and
continues with the current activity of intentional abstract actions
only if it will achieve the desired goal. If, on the other hand, the
logician finds that pursuing the current activity will not achieve the
desired goal, a new activity is computed and implemented.  We use
CR-Prolog to represent and reason with the coarse-resolution and
fine-resolution representations. We use existing implementations of
probabilistic algorithms for executing concrete actions. The following
running example will be used to describe the components of the
architecture, along with differences from prior work.

\begin{figure}[tbh]
  \centering
  \includegraphics[width=0.95\columnwidth]{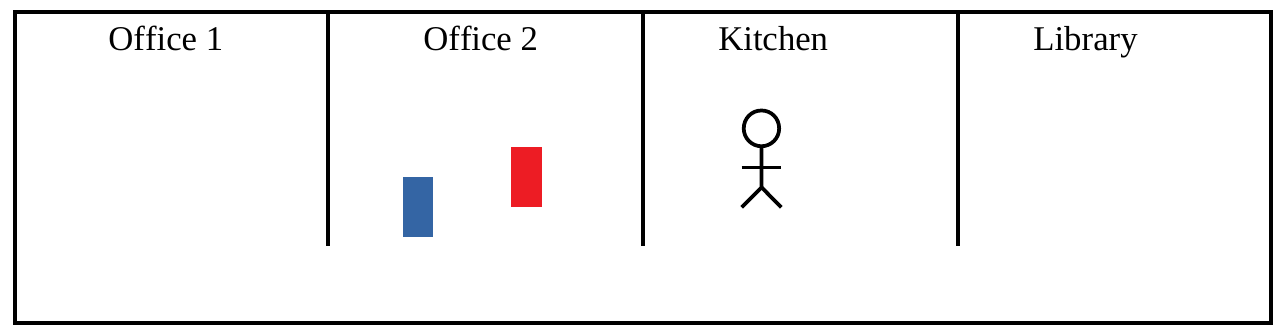}
  \caption{Four rooms considered in Example~\ref{ex:illus-example},
    with a human in the $kitchen$ and two books in $of\!fice_2$.  Only
    the library's door can be locked; all other rooms are open at all
    times.}
  \label{fig:places}
\end{figure}

\begin{example2}\label{ex:illus-example}[Robot Assistant (RA) Domain]
  {\rm Consider a robot assisting humans in moving particular objects
    to desired locations in an indoor office domain with:
    \begin{itemize}
    \item Sorts such as $place$, $thing$, $robot$, $object$, and
      $book$, arranged hierarchically, e.g., $object$ and $robot$ are
      subsorts of $thing$. Sort names and constants are in lower-case,
      and variable names are in uppercase.
    \item Places: $\{of\!fice_1, of\!fice_2, kitchen, library\}$ with
      a door between neighboring places---see Figure~\ref{fig:places};
      only the door between $kitchen$ and $library$ can be locked.
    \item Instances of sorts, e.g., $rob_1$, $book_1$, $book_2$.
    \item Static attributes such as $color$, $size$ and different
      parts (e.g., $base$ and $handle$) associated with objects.
    \item Other agents that may influence the domain, e.g., move a
      book or lock a door. These agents are not modeled explicitly;
      only the potential execution of \emph{exogenous} actions by
      these agents is used to explain unexpected observations.
    \end{itemize}
  }
\end{example2}

\subsection{Action Language and Domain Representation}
\label{sec:arch-ald-represent}
We first describe the action language encoding of the dynamics of the
domain, and the translation of this encoding to CR-Prolog programs for
knowledge representation and reasoning.

\subsubsection{Action Language $\mathcal{AL}_d$}
\label{sec:arch-ald-represent-ald}
Action languages are formal models of parts of natural language used
for describing transition diagrams of dynamic systems. We use an
extension of the action language
$\mathcal{AL}_d$~\cite{gelfond:ANCL13} that supports non-Boolean
fluents and non-deterministic causal laws~\cite{mohan:JAIR19}, to
describe the transition diagrams of our domain at different
resolutions. $\mathcal{AL}_d$ has a sorted signature with
\emph{actions}, i.e., a set of elementary operations, \emph{statics},
i.e., domain attributes whose values cannot be changed by actions, and
\emph{fluents}, i.e., attributes whose values can be changed by
actions. \emph{Basic} fluents obey laws of inertia and can be changed
by actions, whereas \emph{defined} fluents do not obey laws of inertia
and are not changed directly by actions.  $\mathcal{AL}_d$ allows
three types of statements (i) \emph{causal law}; (ii) \emph{state
  constraint}; and (iii) \emph{executability condition}:
\begin{align*}
  a~~\mathbf{causes}~~l_b~~\mathbf{if}~p_0,\ldots,p_m~~~~~~~~&\textrm{(Causal
    law)} \\ \nonumber
  l~~\mathbf{if}~~p_0,\ldots,p_m~~~~~~~~~~~~~~~~~&\textrm{(State
    constraint)}\\\nonumber
  \mathbf{impossible}~a_0,\ldots,a_k~\mathbf{if}~p_0,\ldots,p_m
  ~~~~~~&\textrm{(Executability condition)}
\end{align*}
where $a$ is an action, $l$ is a literal (i.e., a domain attribute or
its negation), $l_b$ is a basic literal, and $p_0,\ldots,p_m$ are
domain literals. The causal law implies that if action $a$ is executed
in a state satisfying $p_0,\ldots,p_m$, the literal $l_b$ will be true
in the resulting state. The state constraint implies that literal $l$
is true in a state satisfying $p_0,\ldots,p_m$. The executability
condition implies that it is impossible to execute actions
$a_0,\ldots,a_k$ in a state satisfying domain literals
$p_0,\ldots,p_m$\footnote{For simplicity, we do not describe the
  non-deterministic causal laws or non-Boolean fluents here.}. For
more details about the syntax and semantics of $\mathcal{AL}_d$,
please see~\cite{gelfond:ANCL13}, and for details about the extension
of $\mathcal{AL}_d$ used in this paper, please
see~\cite{mohan:JAIR19}.

\subsubsection{Coarse-Resolution Knowledge Representation}
\label{sec:arch-ald-represent-coarse}
The coarse-resolution domain representation consists of system
description $\mathcal{D}_c$, which is a collection of statements of
$\mathcal{AL}_d$, and history $\mathcal{H}_c$. System description
$\mathcal{D}_c$ has a sorted signature $\Sigma_c$ and axioms that
describe the corresponding transition diagram $\tau_c$. The signature
$\Sigma_c$ defines the basic sorts, domain attributes and actions. In
addition to the basic sorts and ground instances introduced in
Example~\ref{ex:illus-example}, $\Sigma_c$ for the RA domain includes
sort $step$ for temporal reasoning. Domain attributes (i.e., statics
and fluents) and actions are described in terms of their arguments'
sorts. In the RA domain, coarse-resolution statics include relations
such as $next\_to(place, place)$, which describes the relative
arrangement of places in the domain; and relations modeling object
attributes, e.g., we may represent an object's color as
$obj\_color(object, color)$\footnote{It is possible to represent and
  reason about actions that change the values of the object attributes
  such as color; we just choose to represent them as statics in this
  work.}.  Fluents of the coarse-resolution representation of the RA
domain include $loc(thing, place)$, which denotes the location of the
robot or other domain objects; $in\_hand(robot, object)$, which
denotes whether a particular object is in the robot's hand; and
$locked(place)$, which implies that a particular place is locked. The
locations of other agents, if any, are not changed by the robot's
actions; these locations are inferred from observations obtained from
other sensors.  Next, $\Sigma_c$ for the RA domain includes actions
such as $move(robot, place)$, $pickup(robot, object)$, $putdown(robot,
object)$, and $unlock(robot, place)$; we also consider exogenous
actions $exo\_move(object, place)$ and $exo\_lock(place)$ for
diagnostic reasoning, e.g., for explaining unexpected observations.
Finally, $\Sigma_c$ also includes the relation $holds(fluent, step)$
to imply that a particular fluent is true at a particular time step.
Note that it is possible to consider domain attributes and actions as
functions and use the corresponding notation, e.g., $loc: thing \to
place$, $in\_hand: robot\times object \to bool$, and $move:
robot\times place \to action$. We use the predicate notation for
simplicity, ease of understanding, and to be consistent with the
notation used in other parts of this paper.

Axioms in the coarse-resolution representation of the RA domain
include causal laws, state constraints, and executability conditions
such as:
\begin{subequations}
  \label{eqn:axioms}
  \begin{align}
    &move(rob_1, P)~~\mathbf{causes}~~loc(rob_1, P) \\
    &pickup(rob_1, O)~~\mathbf{causes}~~in\_hand(rob_1, O) \\
    &\neg loc(Th, P_2)~~\mathbf{if}~~loc(Th, P_1),~ P_1\neq P_2 \\
    &loc(O, P)~~\mathbf{if}~~ loc(rob_1, P),~in\_hand(rob_1, O)\\
    &\mathbf{impossible}~~pickup(rob_1, O)~~\mathbf{if}~~loc(rob_1,
    P_1),~loc(O, P_2),~P_1\neq P_2 \\
    &\mathbf{impossible}~~move(rob_1, P)~~\mathbf{if}~~loc(rob_1,
    P_1),~\lnot next\_to(P, P_1)
  \end{align}
\end{subequations}
which describe the dynamics of the domain. For instance,
Statement~\ref{eqn:axioms}(a) implies that executing action
$move(rob_1, library)$ causes $loc(rob_1, library)$ to be true in the
resultant state, Statement~\ref{eqn:axioms}(c) implies that any object
can only be in one location at a time, and
Statement~\ref{eqn:axioms}(e) implies that the robot cannot pick an
object up unless the object is in the same location as the robot.
These axioms are used for inference, planning, and diagnostics, as
described later in Section~\ref{sec:arch-ald-represent-reason}.

The history $\mathcal{H}_c$ of a dynamic domain is usually a record of
statements of the form: (i) $obs(fluent, boolean, step)$ implying that
particular fluents were observed to be true or false at a particular
time step; and (ii) $hpd(action, step)$ implying that particular
actions happened at a particular time step.  In~\cite{mohan:JAIR19},
this notion was expanded to represent defaults describing the values
of fluents in the initial state. For instance, in the
coarse-resolution history $\mathcal{H}_c$ of the RA domain, the
statement ``a book is usually in the library and if it is not there,
it is normally in the office'' is encoded as:
\begin{subequations}
  \begin{align}
    {\bf initial\ default}\ loc(X, library)\ &~{\bf if}~~
    book(X)\\
    {\bf initial~~default}~~loc(X, office_1)&~~{\bf if}~~
    book(X),~~\lnot loc(X, library)
  \end{align}
\end{subequations}
These statements represent prioritized defaults. We can also encode
exceptions, e.g., ``cookbooks are in the kitchen''; for more
information, please see~\cite{gelfond2014knowledge}.  Notice that this
representation does not assign numerical values to degrees of belief
associated with these defaults, but supports elegant reasoning with
generic defaults and their specific exceptions (if any).

\subsubsection{Reasoning with Knowledge}
\label{sec:arch-ald-represent-reason}
Key tasks of an agent equipped with a system description and history
include reasoning with this domain representation for planning and
diagnostics. In our architecture, these tasks are accomplished by
translating the domain representation to a program in CR-Prolog, a
variant of ASP that incorporates consistency restoring (CR)
rules~\cite{balduccini:aaaisymp03}. An independent group of
researchers have developed (and will be releasing) software to
automate the translation between a description in $\mathcal{AL}_d$ and
the corresponding description in CR-Prolog. In our case, we build on
previous work that specified steps for this
translation~\cite{mohan:JAIR19}, and either perform this translation
manually or use a script that automates this translation.

ASP is based on stable model semantics and supports concepts such as
\emph{default negation} and \emph{epistemic disjunction}, e.g., unlike
``$\lnot a$'' that states \emph{a is believed to be false},
``$not~a$'' only implies \emph{a is not believed to be true}, and
unlike ``$p~\lor\,\,\lnot p$'' in propositional logic, ``$p~or~\lnot
p$'' is not tautologous. In other words, each literal can be true,
false or just ``unknown'', and an agent associated with an ASP program
only believes that which it is forced to believe. ASP can also
represent recursive definitions and constructs that are difficult to
express in classical logic formalisms, and it supports non-monotonic
logical reasoning, i.e., it is able to revise previously held
conclusions based on new evidence.  The CR-Prolog program
$\Pi(\mathcal{D}_c, \mathcal{H}_c)$ for the coarse-resolution
representation of the RA domain includes the signature and axioms of
$\mathcal{D}_c$, inertia axioms, reality checks, closed world
assumptions (CWAs) for defined fluents and actions. For instance,
$\Pi(\mathcal{D}_c, \mathcal{H}_c)$ includes:
\begin{subequations}
  \label{eqn:crprolog-axioms}
  \begin{align}
    &holds(F, I+1) \leftarrow~~holds(F, I),~not~\lnot holds(F, I+1) \\
    &\lnot holds(F, I+1) \leftarrow~~\lnot holds(F, I),~not~holds(F, I+1) \\
    &\leftarrow \lnot holds(F, I),~ obs(F, true, I)\\
    &\leftarrow~holds(F, I),~ obs(F, false, I)\\
    & \lnot occurs(A,I) \leftarrow~~not~ occurs(A,I)
  \end{align}
\end{subequations}  
where Statements~\ref{eqn:crprolog-axioms}(a)-(b) are inertia axioms
for basic fluents, Statements~\ref{eqn:crprolog-axioms}(c)-(d) are
reality check axioms implying that any mismatch between observations
and expectations based on current beliefs results in an inconsistency,
and Statement~\ref{eqn:crprolog-axioms}(e) is the CWA for actions.

Program $\Pi(\mathcal{D}_c, \mathcal{H}_c)$ also includes
observations, actions, and defaults from $\mathcal{H}_c$. Every
default also has a CR rule that allows the robot to assume the
default's conclusion is false to restore consistency under exceptional
circumstances. For instance, the axiom:
\begin{align}
  \lnot loc(X, library) \rif book(X)
\end{align}
considers the rare event of a book not being in the library. This
axiom is only used under exceptional circumstances to restore
consistency in the presence of an unexpected observation, e.g., a book
that is expected to be in the $library$ is later found to be in
$of\!\!fice_2$. Each \emph{answer set} of an ASP program, typically
computed by applying a SAT (i.e., satisfiability) solver to the ASP
program, represents the set of beliefs of an agent associated with the
program.  Algorithms for computing entailment, and for planning and
diagnostics, reduce these tasks to computing answer sets of CR-Prolog
programs. We compute answer sets of CR-Prolog programs using the
system called SPARC~\cite{balai:lpnmr13}. An illustrative version of
the coarse-resolution CR-Prolog program for the RA domain (written
using SPARC) is available in our open-source software
repository~\cite{code-results}.

\subsection{Adapted Theory of Intention}
\label{sec:arch-toi}
For any given goal, a robot reasoning with domain knowledge (as
described above) will compute a plan and execute it actions in the
plan until either the goal is achieved or an action in the plan has an
unexpected outcome. In the latter case, the robot will attempt to
explain the unexpected outcome (i.e., perform diagnostics) and compute
a new plan if necessary. To motivate the need for a different approach
in dynamic domains, consider the following five scenarios in which the
goal is to move $book_1$ and $book_2$ to the $library$; these
scenarios have been adapted from scenarios considered in prior
work~\cite{blount2015theory}:
\begin{itemize}
\item \textbf{Scenario 1 (planning):} Robot $rob_1$ is in the kitchen
  holding $book_1$, and believes $book_2$ is in the $kitchen$ and that
  the $library$ is unlocked. The computed plan is:
  \begin{align*}
    &move(rob_1, library),~~putdown(rob_1, book_1)\\
    &move(rob_1, kitchen),~~pickup(rob_1, book_2)\\
    &move(rob_1, library),~~putdown(rob_1, book_2)
  \end{align*}

\item \textbf{Scenario 2 (unexpected success):} Assume that $rob_1$ in
  Scenario-1 has moved to the $library$ and put $book_1$ down, and
  observes $book_2$ there. The robot should be able to explain this
  observation (e.g., $book_2$ was moved there as a result of an
  exogenous action) and realize that the goal has been achieved.

\item \textbf{Scenario 3 (not expected to achieve goal, diagnose and
    replan, case 1):} Assume $rob_1$ in Scenario-1 starts moving
  $book_1$ to $library$, but observes $book_2$ is not in the
  $kitchen$. The robot should realize the plan will fail to achieve
  the overall goal, explain the unexpected observation, and compute a
  new plan.

\item \textbf{Scenario 4 (not expected to achieve goal, diagnose and
    replan, case 2):} Assume $rob1$ is in the kitchen holding $book1$,
  and believes that $book2$ is in $of\!\!fice_2$ and the $library$ is
  unlocked. The robot plans to put $book_1$ in the $library$ before
  fetching $book_2$ from $of\!\!fice_2$. Before $rob_1$ moves to the
  $library$, it unexpectedly observes $book_2$ in the $kitchen$. The
  robot should realize that its current plan will fail, explain the
  unexpected observation, and compute a new plan.

\item \textbf{Scenario 5 (failure to achieve the goal, diagnose and
    replan):} Assume robot $rob_1$ in Scenario-1 is putting $book_2$
  in the $library$, after having put $book_1$ in the $library$
  earlier, and observes that $book_1$ is no longer there. The robot's
  intention should persist; it should explain the unexpected
  observation, replan if necessary, and execute actions until the goal
  is achieved (i.e., both books are in the $library$).

\end{itemize}
One way to support the desired behavior in such scenarios is to reason
with all observations of domain objects and events, e.g., observations
of all objects in the field of view of the robot's (or the domain's)
sensors, during plan execution. Such an approach would be
computationally unfeasible in complex domains in which there may be
many new observations and events at each time step. Also, only a small
number of these observations and events may be relevant to the task at
hand. We thus pursue a different approach in our architecture; our
adapted theory of intention builds on the principles of
non-procrastination and persistence, and extends the ideas from
$\mathcal{TI}$.  Specifically, our architecture enables the robot to
automatically compute actions that are intended for the given goal and
current beliefs. As the robot attempts to implement each such action,
the robot automatically identifies and considers those observations
that are ``relevant'' to this action or the goal. The robot adds these
observations to the recorded history, and uses them to reason about
mental states and actions, to determine if and when it should replan
as against following the existing plan. We will henceforth use
$\mathcal{ATI}$ to refer to this adapted theory of intention; it
expands both the system description $\mathcal{D}_c$ and history
$\mathcal{H}_c$ in the original program $\Pi(\mathcal{D}_c,
\mathcal{H}_c)$ to reason about intentional actions and beliefs.
Below, we describe the steps of this expansion along with some
examples, and provide a link to an illustrative program that is
obtained by applying these steps in the RA domain.

First, the signature $\Sigma_c$ is expanded to represent an
\emph{activity} as a triplet comprising a \emph{goal}, a \emph{plan}
to achieve the goal, and a specific \emph{name} for the activity. We
do so by introducing (in $\Sigma_c$) relations such as:
\begin{align}
  &activity(name),~~activity\_goal(name, goal)\\\nonumber
  &activity\_length(name, length)\\ \nonumber
  &activity\_component(name, number, action)
\end{align}
which represent each named activity, the goal and length of each
activity, and the actions that are the components of the activity.
Note that these relations are not ground initially because the
specific activities and goals are constructed or defined as needed.
However, once they are ground, the corresponding terms behave as
statics.

Next, the existing fluents of $\Sigma_c$ are considered to be
\emph{physical fluents} and the set of fluents is expanded to include
\emph{mental fluents} such as:
\begin{align}
  &active\_activity(activity),~in\_progress\_goal(goal)\\ \nonumber
  &next\_action(activity, action),\\ \nonumber
  &in\_progress\_activity(activity),\\ \nonumber
  &active\_goal(goal),~next\_activity\_name(name)\\ \nonumber
  &current\_action\_index(activity, index)
\end{align}
where the relations in the first three lines are defined fluents,
whereas the other relations are basic fluents that obey the laws of
inertia. All these fluents represent the robot's belief about a
particular activity, action, or goal being active or in progress. None
of these mental fluents' values are changed directly by executing any
physical action. For example, the value of the relation
\emph{current\_action\_index} changes if the robot has completed an
intended action or if a change in the domain makes it impossible for
an activity to succeed. The values of the other mental fluents are
changed directly or indirectly by expanding the set of existing
\emph{physical actions} of $\Sigma_c$ to include \emph{mental actions}
such as:
\begin{align}
  &start(name),~~stop(name)\\ \nonumber
  &select(goal),~~abandon(goal)
\end{align}
where the first two mental actions are used by the controller to start
or stop a particular activity, and the other two actions represent
exogenous actions executed (e.g., by a human or an external system) to
select or abandon a goal.

In addition to $\Sigma_c$, the domain's history $\mathcal{H}_c$ is
expanded to include relations such as:
\begin{align}
  &attempt(action, step)\\ \nonumber
  &\neg~hpd(action, step)
\end{align}
which denote that a particular action was attempted at a particular
time step, and that a particular action did not happen (i.e., was not
executed successfully) at a particular time step. Note that it is
straightforward for the robot to figure out when an action was
attempted, but figuring out when an action was actually executed (or
not executed) requires external (e.g., sensor) input and reasoning,
e.g., diagnostic reasoning with observations to determine whether an
action had the intended outcome(s). In our control loop and
experimental trials, we use ASP to reason with the observations to
determine whether an action was actually executed, and then use this
information for subsequent reasoning.

The expansion of the signature and the history makes it necessary to
expand the description of the domain dynamics. To do so, we introduce
new axioms in $\mathcal{D}'_c$. This includes axioms that represent
the effects of the physical and mental actions on the physical and
mental fluents, e.g., starting (stopping) an activity makes it active
(inactive), and executing an action in an activity keeps the current
activity active.  The new axioms include state constraints, e.g., to
describe conditions under which any particular activity or goal is
active, and executability conditions, e.g., it is not possible for the
robot to simultaneously execute two mental actions or to start an
activity when another activity is active and still valid. In addition,
axioms are introduced to generate intentional actions, build a
consistent model of the domain history, and to perform diagnostics.
For example, the following axioms are related to finding the next
intended action given an activity and a goal:
\begin{subequations}
  \label{eqn:axioms-ati}
  \begin{align}
    &occurs(AA,I1)~~\leftarrow~~ current\_step(I),~I <=
    I1,~\#agent\_action(AA), \\ \nonumber
    &~~~~~~~~~~~~~~~~active\_goal\_activity(AN,I),
    ~holds(in\_progress\_activity(AN),I1), \\ \nonumber
    &~~~~~~~~~~~~~~~holds(next\_action(AN,AA),I1), ~not~impossible(AA,I1) \\ 
    &projected\_success(AN,I)~~\leftarrow~~ current\_step(I),~I < I1,
    ~holds(G, I1)\\\nonumber
    &~~~~~~~~~~~~~~~holds(active\_activity(AN),I1),~activity\_goal(AN,G) \\ 
    &\lnot projected\_success(AN,I)~~\leftarrow~~ current\_step(I),\\\nonumber 
    &~~~~~~~~~~~~~~~not~projected\_success(AN,I)\\ 
    &intended\_action(AA,I)~~\leftarrow~~
    current\_step(I),~\#agent\_action(AA), \\\nonumber 
    &~~~~~~~~~~~~~~~active\_goal\_activity(AN,I),~holds(next\_action(AN,AA),I),
    \\\nonumber 
    &~~~~~~~~~~~~~~~projected\_success(AN,I)
  \end{align}
\end{subequations}
where Statement~\ref{eqn:axioms-ati}(a) implies that if the next agent
action $AA$ in the current activity is not impossible, it is expected
to occur in a subsequent time step. Statement~\ref{eqn:axioms-ati}(b)
implies that if the goal holds in a future step, given that actions of
the current activity $AN$ occur as planned, the activity has a
projected success.  Statement~\ref{eqn:axioms-ati}(c) implies that if
we do not have a projected success, it must have been because one of
our actions cannot occur, or our current activity $AN$ does not reach
the goal.  Finally, Statement~\ref{eqn:axioms-ati}(d) implies that if
our current activity has a projected success, the activity's next
action will be the next intended action.

As another example, the following axioms of $\mathcal{D}'_c$ define an
activity as being futile:
\begin{subequations}
  \label{eqn:axioms-ati-2}
  \begin{align}
    &\leftarrow~ current\_step(I),~active\_goal\_activity(AN,I),\\\nonumber
    &~~~~~~~~~~~\lnot~projected\_success(AN,I),~not~futile\_activity(AN,I) \\ 
    &futile\_activity(AN,I)~\rif~ current\_step(I),
    ~active\_goal\_activity(AN,I), \\ \nonumber
    &~~~~~~~~~~~~~~~~~\lnot~projected\_success(AN,I)\\ \nonumber 
    &intended\_action(stop(AN),I)~\leftarrow~current_step(I),
    ~active\_goal\_activity(AN,I), \\ 
    &~~~~~~~~~~~~~~~~futile\_activity(AN,I) 
  \end{align}
\end{subequations}
where Statement~\ref{eqn:axioms-ati-2}(a) introduces an inconsistency
if there is an active activity $AN$ that does not have projected
success and has not been defined as being futile.  Then,
Statement~\ref{eqn:axioms-ati-2}(b) is a CR rule that provides a path
out of such an inconsistency by defining the activity as being futile
in these exceptional circumstances. Finally,
Statement~\ref{eqn:axioms-ati-2}(c) implies that is an activity as
been defined as being futile, the next intentional action of the robot
will be to stop this activity and plan a different activity.

As described in Section~\ref{sec:arch-ald-represent-reason}, we use a
script to automatically translate the revised system description
$\mathcal{D}'_c$ and history $\mathcal{H}'_c$ to a CR-Prolog program
$\Pi(\mathcal{D}'_c, \mathcal{H}'_c)$ that is solved for planning or
diagnostics. However, recall that CR-rules are used to build a
consistent model of history, which involves reasoning about potential
exceptions to defaults and the execution of exogenous actions, and to
generate minimal plans of intentional actions. This reasoning is
challenging because we need to encode some preferences between the
different CR rules; unexpected observations could potentially be
explained using exceptions to defaults or using exogenous actions,
e.g., a book may be observed in the $kitchen$ because it is an
exception to the corresponding default, or because it was moved there
by some one. Our preference is based on the following key postulate:
assumption:

\medskip
\noindent
\emph{Unexpected observations are more likely to be due to exceptions
  to defaults than due to exogenous actions}.

\medskip
\noindent
This is a reasonable claims for many robotics domains, and is
translated to the following preference:

\medskip
\noindent
\emph{First try to explain unexpected observations by considering
  exceptions to defaults; if that does not suffice, consider exogenous
  actions to generate explanations}.

\medskip
\noindent
Even with this preference, the agent will have to use CR rules for
both diagnostics and planning. Exploring all possible combinations of
such rules can become computationally expensive in complex domains. To
ensure efficient and correct reasoning while still encoding the
desired preference, we modify the axioms (by adding suitable flags)
such that coarse-resolution reasoning with $\mathcal{ATI}$ is
performed in two phases. The robot first computes a consistent model
of history without considering axioms for plan generation, and then
uses this model to guide the computation of plan(s) without
considering the axioms for diagnostics.  A CR-Prolog program
illustrating this process for the RA domain, written in SPARC with
explanatory comments, is available in \emph{ToI\_planning.sp} in the
folder $simulation/ASP\_files/$ in our open-source software
repository~\cite{code-results}.


\medskip
\noindent
The following are the key differences that distinguish $\mathcal{ATI}$
from the prior work on $\mathcal{TI}$ and the prior work on
coarse-resolution reasoning in the refinement-based
architecture~\cite{mohan:JAIR19}::
\begin{enumerate}
\item $\mathcal{TI}$ becomes computationally expensive, especially as
  the size of the plan or domain history increases. Reasoning with
  $\mathcal{TI}$ performs diagnostics and planning jointly, which
  allows it to consider different explanations during planning but
  makes computation unfeasible in all but the very simple domains. On
  the other hand, reasoning with $\mathcal{ATI}$, as stated above,
  first builds a consistent model of history by considering different
  explanations, and \emph{uses the chosen model to guide planning},
  significantly improving computational efficiency and supporting
  scalability in complex domains.

\item $\mathcal{TI}$ assumes complete knowledge of the state of other
  agents (e.g., humans or other robots) that perform exogenous
  actions. In most robotics domains, this assumption is unrealistic;
  these domains typically only afford partial observability.
  $\mathcal{ATI}$ instead makes the more realistic assumption that the
  robot can only make unreliable observations of its domain through
  its sensors and infer exogenous actions by reasoning about and
  trying to explain these observations.

\item $\mathcal{ATI}$ does not include the notion of sub-goals and
  sub-activities (and associated relations) from $\mathcal{TI}$, as
  they are not necessary. Also, these sub-activities and sub-goals
  need to be encoded in advance to use $\mathcal{TI}$, which is
  difficult to do in practical (robotics) domains. Furthermore, even
  if this knowledge is encoded, it will make reasoning (e.g., for
  planning or diagnostics) significantly more computationally
  expensive if the robot has to repeatedly examine if one of the many
  stored activities provides a minimal and correct path to the desired
  goal.

\item Coarse-resolution reasoning in the prior work on the
  refinement-based architecture did not (a) reason about intentional
  actions; or (b) reason about exogenous actions in addition to
  initial state defaults. These limitations are relaxed in the
  architecture described in this paper. A consistent model of history
  is constructed with defaults and exogenous actions at the coarse
  resolution, and reasoning with intentional actions supports
  reasoning in the presence of unexpected successes and failures.
\end{enumerate}
Any architecture with $\mathcal{ATI}$, the original $\mathcal{TI}$, or
a different reasoning component based on non-monotonic logics or
classical first-order logic, will have two key limitations that have
not been discussed so far. First, reasoning does not scale well to the
finer resolution at which actions will often have to be executed to
perform various tasks in robotics domains. For instance, the
coarse-resolution representation discussed so far is not sufficient if
the robot has to grasp and pickup a particular cup from a particular
table, or deliver the cup to a particular person. Also, using logics
to reason with a sufficiently fine-grained domain representation
(e.g., to perform the grasping task) will be computationally
expensive.  Second, we have not yet modeled the actual sensor
observations of the robot or the uncertainty in sensing and actuation.
This uncertainty is the primary source of error on robots, and many
existing algorithms use probabilities to model this uncertainty
quantitatively. Section~\ref{sec:relwork} discusses additional
limitations of approaches based on logical and probabilistic reasoning
for robotics domains. Our architecture addresses these limitations by
combining $\mathcal{ATI}$ with ideas that build on prior work on a
refinement-based architecture~\cite{mohan:JAIR19}, as described below.


\subsection{Refinement, Zooming and Execution}
\label{sec:arch-refine-zoom}
Consider a coarse-resolution system description $\mathcal{D}_c$ of
transition diagram $\tau_c$ that includes $\mathcal{ATI}$. For any
given goal, reasoning with $\Pi(\mathcal{D}_c, \mathcal{H}_c)$ will
provide an activity, i.e., a sequence of abstract intentional actions.
In our architecture, the execution of the coarse-resolution transition
corresponding to each such abstract action is based on a
fine-resolution system description $\mathcal{D}_f$ of transition
diagram $\tau_f$ that is a \emph{refinement} of, and is tightly
coupled to, $\mathcal{D}_c$. We can imagine refinement as taking a
closer look at the domain through a magnifying lens, potentially
leading to the discovery of concrete structures that were previously
abstracted away by the designer, e.g., for efficient reasoning with
rich commonsense knowledge. Our architecture builds on the general
design methodology described in prior work~\cite{mohan:JAIR19} to
construct $\mathcal{D}_f$ using $\mathcal{D}_c$ and some
domain-specific information provided by the designer. This approach
includes a weak refinement that temporarily limits the robot's ability
to observe the value of fluents (through sensors), and a theory of
observation that leads to the definition of strong refinement by
relaxing this limitation. The coarse-resolution transition is then
implemented by automatically \emph{zooming} to and reasoning with the
part of $\mathcal{D}_f$ relevant to this transition and the
coarse-resolution goal. We describe the steps of this process and
highlight key differences between our current approach and prior
work~\cite{mohan:JAIR19}.

First, the signature $\Sigma_f$ of $\mathcal{D}_f$ includes all
elements of $\Sigma_c$ except those related to $\mathcal{ATI}$, and a
new sort for every sort of $\Sigma_c$ that is \emph{magnified} by the
increase in resolution; these new sorts are the fine-resolution
\emph{counterparts} of the magnified sorts.  For instance, $\Sigma_f$
of the RA domain includes:
\begin{align}
  &place=\{office_1, office_2, kitchen, library\},~~cup=\{cup_1\},
  \\\nonumber &~~place^*=\{c_1,\dots,c_m\},~~cup^*=\{cup_1\_base,
  cup_1\_handle\}\\ \nonumber &book=\{book_1, book_2\}
\end{align}
where the superscript ``*'' represents fine-resolution counterparts of
the sorts in $\mathcal{D}_c$ that are magnified by refinement.  Also,
$\{c_1,\dots,c_m\}$ are the grid cells that are the components of the
original set of places, and any $cup$ has a $base$ and $handle$ as
components (i.e., parts); a $book$, on the other hand, is not
magnified and has no components. The sort hierarchy is also suitably
modified, e.g., $cup$ and $cup^*$ are siblings that are children of
sort $object$. Also, for each domain attribute of $\Sigma_c$ magnified
by the increase in resolution, we introduce appropriate
fine-resolution counterparts in $\Sigma_f$. For instance, in the RA
domain, $\Sigma_f$ includes domain attributes such as:
\begin{align}
  &loc(thing, place),~next\_to(place, place)\\ \nonumber
  &loc^*(thing^*, place^*),~next\_to^*(place^*, place^*)
\end{align}
where relations with and without the ``*'' superscript represent the
fine-resolution counterparts and their coarse-resolution versions
respectively. The specific relations listed above describe the
location of each thing at two different resolutions, and describe two
places or cells that are next to each other. The signature $\Sigma_f$
will also include actions that are copies of those in $\Sigma_c$ and
those with magnified sorts. For instance, $\Sigma_f$ for the RA domain
includes:
\begin{align}
  &move(robot, place), ~~in\_hand(robot, object)\\ \nonumber
  &move^*(robot, place^*), ~~in\_hand^*(robot, cup^*)
\end{align}
Finally, the signature $\Sigma_f$ includes domain-dependent statics
$component(O^*, O)$ relating the magnified objects and their
counterparts, e.g., $component(cup_1\_base, cup_1)$ describes that the
base of a cup is a component of the corresponding cup. The axioms of
$\mathcal{D}_f$ are then obtained by restricting the axioms of
$\mathcal{D}_c$ (except those for $\mathcal{ATI}$) to the signature
$\Sigma_f$. This would, for instance, remove all axioms related to
$\mathcal{ATI}$, leave certain axioms unchanged, and introduce
fine-resolution versions of certain axioms.  For instance, the axioms
in $\mathcal{D}_f$ for the RA domain will now include:
\begin{subequations}
  \begin{align}
    &move^*(R, C)~~\mathbf{causes}~~loc^*(R, C)\\
    &pickup(R, O)~~\mathbf{causes}~~in\_hand(R, O)\\
    &pickup^*(R, Opart)~~\mathbf{causes}~~in\_hand^*(R, Opart)\\
    &next\_to^*(C_2, C_1)~~\mathbf{if}~~next\_to^*(C_1, C_2)
  \end{align}
\end{subequations}
where $C$, $C_1$ and $C_2$ are elements of sort $place^*$ (i.e., grid
cells in places), and $Opart$ is an element of sort $cup^*$, i.e.,
$cup_1\_base$ or $cup_1\_handle$. We also include bridge axioms that
relate coarse-resolution domain attributes to their fine-resolution
counterparts. For instance:
\begin{subequations}
  \label{eqn:bridge-axioms}
  \begin{align}
    &loc(O, P)~~\mathbf{if}~~component(C, P),~loc^*(O, C)\\
    &in\_hand(R, O)~~\mathbf{if}~~component(Opart, O), ~in\_hand^*(R, Opart)
  \end{align}
\end{subequations}
where Statement~\ref{eqn:bridge-axioms}(a) implies that any object
that is in a particular cell within a particular room is also within
that room, and Statement~\ref{eqn:bridge-axioms}(b) implies that if
the robot has some part of an object in its grasp then the entire
object is also in its grasp. Note that the refinement process does not
inherit any of the relations or axioms that were introduced in
$\mathcal{D}_c$ to reason about intentional actions.

Next, to support the observation of the values of fluents, the
signature $\Sigma_f$ is expanded to include \emph{knowledge-producing}
action $test(robot, fluent)$ that checks the value of a fluent in a
given state, and only changes the value of appropriate
(fine-resolution) knowledge fluents.  We also introduce
\emph{knowledge fluents} to describe observations of the environment,
e.g., basic fluents to describe the direct (sensor-based) observation
of the values of the fine-resolution fluents, and defined
domain-dependent fluents that determine when the value of a particular
fluent can be tested. Note that the value of any concrete fluent or
static in $\Sigma_f$ is directly observable, e.g., the grid cell
location of the robot, whereas any abstract fluent or static in
$\Sigma_f$ is only indirectly observable, e.g., the place location of
an object cannot be observed directly. The axioms of $\mathcal{D}_f$
are then expanded to include (a) causal laws describing the effect of
the $test$ action on the corresponding fine-resolution basic knowledge
fluents; (b) executability conditions for these $test$ actions; (c)
axioms that describe the robot's ability to sense the values of
directly and indirectly observable fluents; and (d) auxiliary axioms
for indirect observation of fluents. For example:
\begin{subequations}
  \label{eqn:tobs-axioms}
  \begin{align}
    &test(rob_1, F) ~{\bf causes}~~observed(rob_1,F)~~{\bf if }~~F=true \\
    &{\bf impossible}~~test(rob_1, F)~~{\bf if}~~\lnot can\_test(rob_1, F) \\
    &can\_test(rob_1, in\_hand(rob_1, O)) \\
    &observed(rob_1, loc(O, P))~~\mathbf{if}~~observed(rob_1, loc(O,
    C)), ~component(C, P)
  \end{align}
\end{subequations}
where $can\_test(rob_1, F)$ is a domain-dependent defined fluent that
encodes the information about when the robot can test the value of a
particular fluent, and $observed(rob_1, F)$ is a knowledge fluent that
encodes that the robot has observed a particular value for a
particular fluent directly, e.g., Statement~\ref{eqn:tobs-axioms}(a),
or indirectly, e.g., Statement~\ref{eqn:tobs-axioms}(d). Prior work
has shown that if certain conditions are met by the definition of
$\mathcal{D}_f$ and $\mathcal{D}_c$, then for each transition in
$\tau_c$ between coarse-resolution states $\sigma_1$ and $\sigma_2$,
there exists a path in $\tau_f$ between some refinement of $\sigma_1$
and some refinement of $\sigma_2$---see~\cite{mohan:JAIR19} for
related definitions and proofs. Although the $\mathcal{D}_c$ described
in this paper also includes $\mathcal{ATI}$, recall that the design of
our architecture includes the key decision of confining the
representation and reasoning methods associated with this theory to
the coarse resolution. In other words, although the transition
diagrams for $\mathcal{D}_c$ and $\mathcal{D}_f$, i.e., $\tau_c$ and
$\tau_f$, are tightly-coupled, the components of the signature and the
axioms added to $\mathcal{D}_c$ for $\mathcal{ATI}$ are not refined or
included in $\mathcal{D}_f$. Our design choice thus enables us to
include the additional theory while ensuring that the result
from~\cite{mohan:JAIR19} about the correspondence of paths in $\tau_c$
and $\tau_f$ holds for the coarse and fine resolution descriptions in
this paper.

While the tight coupling established by refinement between the coarse
resolution and fine resolution descriptions is appealing, reasoning at
fine resolution using $\mathcal{D}_f$ becomes computationally
unfeasible for complex domains. Also, the refined description does not
(so far) consider probabilistic models of the uncertainty in sensing
and actuation. We address the computational complexity problem through
a key expansion to the principle of \emph{zooming} introduced
in~\cite{mohan:JAIR19}.  Specifically, for each abstract transition
$T$ to be implemented (i.e., executed) at fine resolution, the
previous definition of zooming determined $\mathcal{D}_f(T)$, the part
of the system description $\mathcal{D}_f$ relevant to transition $T$;
it did so by determining the object constants of $\Sigma_f$ relevant
to $T$ and restricting $\mathcal{D}_f$ to these object constants.
Here, we extend this definition of zooming to identify
$\mathcal{D}_f(T, G)$, the part of system description $\mathcal{D}_f$
relevant to the transition or the overall goal. To identify this part,
we first make some key changes to the definition of relevance
in~\cite{mohan:JAIR19} as follows.

\medskip
\begin{definition}\label{def:relevant}[Constants relevant to a transition or goal]\\
  {\rm For any given (ground) transition $T = \langle \sigma_1, a^H,
    \sigma_2\rangle$ of $\tau_c$ and goal $G$, $relCon_c(T,G)$ denotes
    the minimal set of object constants of signature $\Sigma_c$ of
    system description $\mathcal{D}_c$ closed under the following
    rules:
    \begin{enumerate}
    \item Object constants occurring in $a^H$ are in $relCon_c(T,G)$;
    \item Object constants occurring in $G$ are in $relCon_c(T,G)$;
    \item If the term $f(x_1,\ldots, x_n, y)$ belongs to state
      $\sigma_1$ or $\sigma_2$, but not both, then the constants $x_1,
      \ldots, x_n, y$ are in $relCon_c(T,G)$;
    \item If the body of an executability condition of $a^H$ contains
      a term $f(x_1,\dots,x_n, y)$ that is in $\sigma_1$, the
      constants $~x_1,\dots,x_n,y~$ are in $relCon_c(T,G)$;
    \item If $f(x_1,\ldots, x_n, y)$ belongs to $G$, then $x_1,
      \ldots, x_n, y$ are in $relObCon_c(T, G)$.
    \end{enumerate}
    Constants from $relCon_c(T, G)$ are said to be \emph{relevant} to
    $T$ or $G$. }
\end{definition}
\noindent
Note that unlike prior work, this definition of relevance considers
the coarse-resolution goal when identifying the object constants
relevant to a particular coarse-resolution transition. Consider a
scenario in our RA domain, in which the goal is to take the book
$tb_1$, which is known to be in $of\!\!fice_1$, to the $library$, with
the robot $rob_1$ being in the $kitchen$. For the first transition $T
= \langle \sigma_1, move(rob_1, of\!\!fice_1), \sigma_2\rangle$ in the
activity for this goal, $rob_1$ and $tb_1$ are relevant; other domain
objects are not considered to be relevant. Also note that in the
absence of considering object constants relevant to the goal, $tb_1$
would not be in $relCon_c(T, G)$. Now, the coarse-resolution system
description $\mathcal{D}_c(T, G)$ relevant to $T$ or $G$ is obtained
by first constructing signature $\Sigma_c(T, G)$ whose object
constants are those of $relCon_c(T, G)$. Basic sorts of $\Sigma_c(T,
G)$ are intersections of basic sorts of $\Sigma_c$ with those of
$relCon_c(T, G)$, e.g., we would not consider sorts such as $cup$. The
domain attributes and actions of $\Sigma_c(T, G)$ are those of
$\Sigma_c$ restricted to the basic sorts of $\Sigma_c(T, G)$, i.e., we
only retain the domain attributes and actions that can be defined
entirely in terms of the basic sorts of $\Sigma_c(T, G)$; we would
not, for instance, consider $in\_hand(robot, cup)$ or $pickup(robot,
cup)$. In a similar manner, the axioms of $\mathcal{D}_c(T, G)$ are
restrictions of axioms of $\mathcal{D}_c$ to $\Sigma_c(T, G)$, e.g.,
we would not consider axioms related to action $pickup(robot, cup)$.
It is easy to show that for any coarse-resolution transition $T =
\langle \sigma_1, a^H, \sigma_2\rangle$, there exists a transition
$\langle \sigma_1(T, G), a^H, \sigma_2(T, G)\rangle$ in transition
diagram of $\mathcal{D}_c(T, G)$, with $\sigma_1(T, G)$ and
$\sigma_2(T, G)$ being restrictions of $\sigma_1$ and $\sigma_2$
(respectively) to the relevant signature $\Sigma_c(T, G)$.

Once the relevant coarse-resolution system description has been
identified, the zoomed system description can be constructed as
follows.
\medskip
\begin{definition}\label{def:zoomed-description}[Zoomed system description]\\
  {\rm For a coarse-resolution transition $T$ and goal $G$, the
    fine-resolution system description $\mathcal{D}_f(T, G)$ with
    signature $\Sigma_f(T, G)$ is the \emph{zoomed fine-resolution
      system description} if:
    \begin{enumerate}
    \item Basic sorts of $\Sigma_f(T, G)$ are those of
      $\mathcal{D}_f$ that are components of the basic sorts of
      $\mathcal{D}_c(T)$.
    
    \item Functions of $\Sigma_f(T, G)$ are those of $\mathcal{D}_f$
      restricted to the basic sorts of $\Sigma_f(T, G)$.

    \item Actions of $\Sigma_f(T, G)$ are those of $\mathcal{D}_f$
      restricted to the basic sorts of $\Sigma_f(T, G)$.

    \item Axioms of $\mathcal{D}_f(T, G)$ are those of $\mathcal{D}_f$
      restricted to the signature $\Sigma_f(T, G)$.
    \end{enumerate}
  }
\end{definition}
\noindent
Continuing with our example in the RA domain, the object constants of
$\Sigma_f(T, G)$ include $rob_1$, places $\{of\!\!fice_1, kitchen\}$,
cells $\{c_i : c_i\in kitchen\cup of\!\!fice_1\}$, and book $tb_1$.
Domain attributes of $\Sigma_f(T, G)$ include $loc(rob_1, P)$,
$loc^*(rob_1, C)$, $next\_to(P_1, P_2)$, $next\_to^*(C_1, C_2)$, with
the variables taking values from the set of places and cells in
$\Sigma_f(T, G)$, and properly restricted relations for testing and
observing the values of fluents etc. In a similar manner, actions of
$\Sigma_f(T, G)$ include $move^*(rob_1, c_i)$, which moves the robot
to a particular cell, $test(rob_1, loc^*(rob_1, c_i))$, which checks
whether $rob_1$ is in a particular cell location, and $observed(rob_1,
loc(tb_1, c_j))$, which represents the observation of $tb_1$ in a
particular cell. Also, restricting axioms of $\mathcal{D}_f$ to the
signature $\Sigma_f(T, G)$ removes causal laws for $pickup$ and
$putdown$, and irrelevant state constraints and executability
conditions; the variables in the remaining axioms are restricted to
object constants in $\Sigma_f(T, G)$. It can be shown that for any
given transition $\langle \sigma_1(T, G), a^H, \sigma_2(T, G)\rangle$
in the coarse-resolution transition diagram of $\mathcal{D}_c(T, G)$,
there exists a path between a refinement of $\sigma_1(T, G)$ and a
refinement of $\sigma_2(T, G)$. This result can be established by
following steps similar to those in the proof provided in prior
work~\cite{mohan:JAIR19}. The key differences are the revised
definitions of relevance and zooming, as provided above, which will
require suitable revisions in the proof.

Once the relevant fine-resolution description has been identified,
prior work achieved fine-resolution implementations of any desired
coarse resolution transition $T$ by (a) mapping $\mathcal{D}_f(T)$ and
estimated probabilities of state transitions to a partially observable
Markov decision process (POMDP); and (b) using an approximate solver
to solve each such POMDP and obtain a policy that maps belief states
to actions. Although the POMDP that is constructed and solved only
focuses on the relevant part of the fine-resolution description, this
approach can become computationally expensive in complex domains.
Instead, to implement transition $T$ in our architecture, ASP-based
reasoning with $\Pi(\mathcal{D}_f(T, G), \mathcal{H}_f)$ is used to
compute a sequence of concrete (i.e., fine-resolution) actions, with
the goal being a fine-resolution counterpart of the resultant state of
the coarse-resolution transition $T$. In what follows, we use
``refinement and zooming'' to refer to the use of both refinement and
zooming as described above. The execution of each fine-resolution
concrete action is then based on existing implementations of
algorithms for common robotics tasks such as navigation, mapping,
object recognition, localization, and grasping---see
Section~\ref{sec:expres} for more details. These algorithms provide
probabilistic measures of certainty about their decisions, e.g., about
the presence or absence of target objects in an image of the scene.
When the robot makes decisions at the fine resolution, the
high-probability outcomes of each concrete action's execution get
elevated to statements associated with complete certainty in
$\mathcal{H}_f$ and used for subsequent reasoning; this approach may
result in incorrect commitments but the non-monotonic logical
reasoning capability helps the robot identify and recover from such
errors. The coarse-resolution outcomes of such fine-resolution
reasoning are added to the coarse-resolution $\mathcal{H}_c$ for
subsequent reasoning using $\mathcal{ATI}$. The CR-Prolog programs for
fine-resolution reasoning in the RA domain (i.e., with the refined and
zoomed system description), and the program for the overall control
loop of the architecture, are available in our online
repository~\cite{code-results}. 

The following are the key differences that distinguish fine-resolution
reasoning in our architecture from that in prior work on the
refinement-based architecture~\cite{mohan:JAIR19}:
\begin{enumerate}
\item Prior work did not maintain a history and perform logical
  reasoning at the fine-resolution; as stated earlier, a POMDP-based
  approach was used, which becomes computationally expensive in
  complex domains. Also, prior work assumed limited dynamic changes in
  the domain during planning and execution. These limitations are
  relaxed in the architecture described in this paper.
  Fine-resolution reasoning builds a consistent model of history, and
  considers the relevant fine-resolution observations to compute and
  add appropriate statements to the coarse-resolution history.
  Furthermore, the tight coupling between the system descriptions and
  the separation of concerns, with $\mathcal{ATI}$ only included in
  the coarse resolution, helps establish the desirable properties of
  prior work, e.g., about the existence of paths in the
  fine-resolution transition diagram for any given transition in the
  coarse-resolution diagram.

\item Zooming is a key requirement for the desired reasoning
  capabilities and for computational efficiency.  Prior work on
  zooming automatically extracted the part of the fine-resolution
  system description relevant to the implementation of any given
  transition at the coarse resolution. The architecture described in
  this paper, on the other hand, automatically identifies and reasons
  about the part of the fine-resolution system description relevant to
  the coarse-resolution transition and the goal under consideration.
  As a result, reasoning and plan execution are reliable and efficient
  in the presence of dynamic (and unexpected) changes in the domain.


\item Prior work used a POMDP to reason probabilistically over the
  zoomed fine-resolution system description $\mathcal{D}_f(T)$ for any
  coarse-resolution transition $T$. This is a computationally
  expensive process, especially when domain changes prevent reuse of
  POMDP policies~\cite{mohan:JAIR19}. In this paper, CR-Prolog is used
  to compute a plan of concrete actions from $\mathcal{D}_f(T, G)$.
  Each concrete action is then executed using algorithms that
  incorporate probabilistic models of uncertainty, significantly
  reducing the computational cost of fine-resolution reasoning and
  execution. In addition, the algorithms for the individual concrete
  actions can be implemented, revised, and replaced without requiring
  any further changes in the other components of the architecture.
\end{enumerate}
As we show below, these differences help improve the reliability and
computational efficiency of reasoning.

\section{Experimental Setup and Results}
\label{sec:expres}
This section reports the results of experimentally evaluating the
capabilities of our architecture in different scenarios. We evaluated
the following hypotheses:
\begin{itemize}
\item \underline{\textbf{H1:}} using $\mathcal{ATI}$ improves the
  computational efficiency in comparison with not using it, especially
  in scenarios with unexpected success.
\item \underline{\textbf{H2:}} using $\mathcal{ATI}$ improves the
  accuracy in comparison with not using it, especially in scenarios
  with unexpected goal-relevant observations.
\item \underline{\textbf{H3:}} the architecture that combines
  $\mathcal{ATI}$ with refinement and zooming supports reliable and
  efficient operation in complex (robot) domains.
\end{itemize}
We report results of evaluating these hypotheses experimentally: (a)
in a simulated domain based on Example~\ref{ex:illus-example}; (b) on
a Baxter robot manipulating objects on a tabletop; and (c) on a
Turtlebot finding and moving objects to particular places in an indoor
domain. We also provide some execution traces as illustrative examples
of the working of the architecture. To evaluate the ability to scale
to more complex domains, we defined variants of the RA domain at eight
different complexity levels. The key components of each complexity
level are as follows:
\begin{itemize}
\item \underline{\textbf{L1:}} one object with one fine-resolution
  part, i.e., no new parts considered after refinement; two rooms with
  two cells in each room.

\item \underline{\textbf{L2:}} two objects, each with two refined
  parts; three rooms with two cells in each room.

\item \underline{\textbf{L3:}} three objects, each with three
  fine-resolution parts (e.g., base and handle of cup); four rooms
  with four cells in each room.

\item \underline{\textbf{L4:}} four objects, each with four refined
  parts; five rooms with five cells in each room.

\item \underline{\textbf{L5:}} eight objects, each with two refined
  parts; five rooms with nine cells in each room.

\item \underline{\textbf{L6:}} eight objects, each with two
  fine-resolution parts, and four objects, each with one
  fine-resolution part; five rooms with twelve cells in each room.

\item \underline{\textbf{L7:}} eight objects, each with two
  fine-resolution parts, and four objects, each with one
  fine-resolution part; five rooms with sixteen cells in each room.

\item \underline{\textbf{L8:}} sixteen objects, each with two
  fine-resolution parts, and eight objects, each with one
  fine-resolution part; five rooms with sixteen cells in each room.
\end{itemize}
where the number of objects, number of object parts, number of rooms,
and the number of cells in each room, typically increase between
consecutive complexity levels. There are some exceptions, e.g.,
between $L5-L6$ and $L6-L7$, introduced to isolate and study the
effects of a change in the value of one of these parameters.

In each experimental trial, the robot's goal was to find and move one
or more objects to particular locations. As a baseline for comparison
for hypotheses $H1$ and $H2$, we used an ASP-based reasoner that does
not include $\mathcal{ATI}$---we refer to this as the ``traditional
planning'' ($\mathcal{TP}$) approach. The term ``traditional'' implies
that the planner only monitors the effects of the action being
executed; it does not identify and monitor observations related to the
current transition and the goal. We do not use $\mathcal{TI}$ as the
baseline for comparison because it includes components that make it
much more computationally expensive than $\mathcal{ATI}$. Also,
$\mathcal{TI}$ does not support reasoning with incomplete knowledge,
non-determinism, and partial observability, capabilities that are
often needed in robotics domains---see Section~\ref{sec:arch-toi} for
a related discussion. In the $\mathcal{TP}$ approach, the robot uses
ASP to reason with incomplete domain knowledge, and only monitors the
outcome(s) of the action currently being executed.  Recall that
$\mathcal{ATI}$ is introduced in the coarse resolution; to thoroughly
examine the effect of this theory, we first compare $\mathcal{ATI}$
with $\mathcal{TP}$ in the coarse resolution, i.e., without any
refinement, zooming, or fine-resolution reasoning. We then separately
examine the effect of refinement, zooming, and probabilistic models of
the uncertainty in sensing and actuation, in the context of evaluating
hypothesis $H3$.  We do so by combining refinement and zooming with
$\mathcal{ATI}$; the baseline for comparison was a system that did not
use zooming as part of fine-resolution reasoning---we refer to this as
the ``non-zooming'' approach that still includes $\mathcal{ATI}$ (at
the coarse resolution) and reasoning with the refined description. We
also combine $\mathcal{ATI}$ with refinement and zooming to run
experiments on robots. Although we do not do so in this paper, our
architecture's components for fine-resolution reasoning can also be
combined with $\mathcal{TP}$ (if needed).

As stated in Section~\ref{sec:arch-refine-zoom}, we use existing
implementations of suitable algorithms for executing the concrete
actions, e.g., for navigation, object recognition, obstacle avoidance,
and manipulation. These algorithms internally model and estimate the
uncertainty in sensing and actuation probabilistically. Some of these
algorithms operate continuously (e.g., for obstacle avoidance), while
others (e.g., object recognition) are selected and used as needed.
When we run experiments in simulation (see
Section~\ref{sec:expres-sim} below), we use statistics obtained from
executing the concrete actions on robots to simulate the probabilistic
models of uncertainty, e.g., the robot moves to the desired grid cell
in $85\%$ of the trial and recognizes an object correctly in $90\%$ of
the trials. When we run experiments on robots (see
Section~\ref{sec:expres-robots} below), we use existing
implementations of algorithms developed by us and other researchers
based on the Robot Operating System (ROS). For instance, whenever we
use our architecture in a domain where the robot can move, we use the
particle filter-based algorithm in ROS for simultaneous localization
and mapping~\cite{dissanayake:TRA01}.  This algorithm enables the
robot to periodically, simultaneously, and probabilistically track
multiple hypotheses, each of which represent a pose sequence and a map
of the domain. For visual object recognition, we use an algorithm
developed by others in our research group. This algorithm is used when
needed by executing a suitable knowledge-producing (e.g., $test$)
action, and is based on learned models that characterize each object
using color, shape, and local gradient features~\cite{li:icar13}. We
also use an existing implementation in ROS for local obstacle
avoidance. These algorithms associate probabilities with outcomes,
e.g., a probabilistic measure of certainty is computed and provided
with the robot's estimate of its pose, or its estimate of the class
label assigned to domain objects observed in camera images.

We used one or more of the following performance measures to evaluate
the hypotheses: (i) total (planning and execution) time; (ii) number
of plans computed; (iii) planning time; (iv) execution time; (v)
number of actions executed; and (vi) accuracy. Note a plan is
considered to be correct if it is minimal and results (on execution)
in the achievement of the goal. We begin with execution traces
demonstrating the working of the architecture.

\subsection{Execution traces}
\label{sec:expres-traces}
The following execution traces illustrate the differences in the
decisions made by a robot using $\mathcal{ATI}$ in comparison with a
robot using $\mathcal{TP}$, focusing primarily on coarse-resolution
reasoning. These traces correspond to scenarios drawn from the RA
domain; we focus on scenarios in which the robot has to respond to
unexpected observed effects (successes and failures) caused by
exogenous actions.

\begin{execexample}\label{exec:example1}[Example of Scenario-2]\\
  {\rm Assume that robot $rob_1$ is in the $kitchen$ initially,
    holding $book_1$ in its hand, and believes that $book_2$ is in
    $of\!\!fice_2$ and the $library$ is unlocked.
    \begin{itemize}
    \item The goal is to have $book_1$ and $book_2$ in the $library$.
      The computed plan is the same for $\mathcal{ATI}$ and
      $\mathcal{TP}$, and consists of actions:
      \begin{align*}
        &move(rob_1, library),~putdown(rob_1, book_1),\\
        &move(rob_1, kitchen), ~move(rob_1, office_2),\\
        &pickup(rob_1, book_2), ~move(rob_1, kitchen)\\
        &move(rob_1, library), ~putdown(rob_1, book_2)
      \end{align*}
      
    \item Assume that as the robot is putting $book_1$ down in the
      $library$, $book_2$ has been moved (e.g., by a human or other
      external agent) to the $library$.

    \item With $\mathcal{ATI}$, the robot observes $book_2$ in the
      $library$, reasons and explains the observation as the result of
      an exogenous action, realizes the goal has been achieved and
      stops further planning and execution.

    \item With $\mathcal{TP}$, the robot does not observe or does not
      use the information encoded in the observation of $book_2$. It
      will thus waste time executing subsequent steps of the plan
      until it is unable to find or pickup $book_2$ in the $library$.
      It will then replan (potentially including prior observation of
      $book_2$) and eventually achieve the desired goal. It may also
      compute and pursue plans assuming $book_2$ is in different
      places, and take more time to achieve the goal.
    \end{itemize}
  }
\end{execexample}

\begin{execexample}\label{exec:example2}[Example of Scenario-5]\\
  {\rm Assume that robot $rob_1$ is in the $kitchen$ initially,
    holding $book_1$ in its hand, and believes that $book_2$ is in
    $kitchen$ and the $library$ is unlocked.
    \begin{itemize}
    \item The goal is to have $book_1$ and $book_2$ in the $library$.
      The computed plan is the same for $\mathcal{ATI}$ and
      $\mathcal{TP}$, and consists of the actions:
      \begin{align*}
        &move(rob_1, library),~putdown(rob_1, book_1),\\
        &move(rob_1, kitchen), pickup(rob_1, book_2),\\
        &move(rob_1, library), ~putdown(rob_1, book_2)
      \end{align*}
    
    \item Assume the robot is in the act of putting $book_2$ in the
      $library$, after having already put down $book_1$ in the
      $library$ earlier. However, $book_1$ is unexpectedly moved from
      the $library$ (e.g., to the $kitchen$, unknown to the robot)
      while the robot is moving $book_2$.

    \item With $\mathcal{ATI}$, the robot observes $book_1$ in not in
      the $library$, realizes the goal has not been achieved although
      the computed plan has been completed, computes a new plan, and
      executes this plan until it finds $book_1$ and moves it to the
      $library$.

    \item With $\mathcal{TP}$, the robot puts $book_2$ in the
      $library$ and stops execution because it believes it has
      achieved the desired goal. In other words, it does not realize
      that the goal has not been achieved.
    \end{itemize}
  }
\end{execexample}

\subsection{Experimental Results in Simulation}
\label{sec:expres-sim}
We evaluated hypotheses $H1$ and $H2$ extensively in a simulated world
that mimics Example~\ref{ex:illus-example}, with four places and
different objects. Please note the following:
\begin{itemize}
\item As stated earlier, we first compared $\mathcal{ATI}$ with
  $\mathcal{TP}$ in the context of the coarse-resolution domain
  representation, i.e., these trials did not include refinement,
  zooming or fine-resolution reasoning. We also temporarily abstracted
  away uncertainty in perception and actuation.

\item We conducted paired trials and compared the results obtained
  using $\mathcal{TP}$ with those obtained using $\mathcal{ATI}$ for
  the same initial conditions and for the same dynamic domain changes
  (when appropriate), e.g., a book is moved unknown to the robot and
  the robot obtains an unexpected observation.

\item When we included fine-resolution reasoning in simulation, we
  assumed a fixed execution time for each concrete action to measure
  execution time, e.g., $15$ units for moving from a room to the
  neighboring room, $5$ units to pick up an object or put it down; and
  $5$ units to open a door. 

\item Ground truth (e.g., minimal plan) was provided by a separate
  component that reasons with complete domain knowledge.
\end{itemize}
Table~\ref{tab:sim-results} summarizes the results of $\approx 800$
paired trials in each of the five scenarios described in
Section~\ref{sec:arch-toi}. Also, all claims made below were tested
for statistical significance. The initial conditions, e.g., starting
location of the robot and objects' locations, and the goal, were set
randomly in each paired trial. However, before choosing a particular
instance of a scenario defined by a particular initial condition, the
simulator does use ground truth knowledge (not available to the robot)
to verify that the chosen goal is reachable from the chosen initial
conditions. Also, in suitable scenarios, a randomly-chosen, valid
(unexpected) domain change is introduced in each paired trial. Given
the significant differences that may exist between two paired trials,
averaging the measured time or plan length across different trials
does not provide any useful information about the performance of the
two approaches being compared. In each paired trial, the value of each
performance measure (except accuracy) obtained with $\mathcal{TP}$ is
thus expressed as a fraction of the value of the same performance
measure obtained with $\mathcal{ATI}$; each value reported in
Table~\ref{tab:sim-results} is the average of these computed ratios.
We highlight some key findings below.

\begin{table*}[t]
\centering
\small
\setlength\arrayrulewidth{0.5pt}
\begin{tabular}{|c|c|c|c|c|c|c|c|}
  \hline
  \multirow{2}{*}{\textbf{Scenarios}}&  \multicolumn{5}{|c|}{\textbf{Average Ratios}} & \multicolumn{2}{|c|}{\textbf{Accuracy}}\\
  \cline{2-8}
  &\textbf{Tot. Time} &\textbf{\# Plans} & \textbf{Plan. Time} & \textbf{Exec. Time} & \textbf{\# Actions} & $\mathcal{TP}$ & $\mathcal{ATI}$ \\
  \hline
  1 &0.81 &1.00 &0.45  &1.00 &1.00 &100\% &100\% \\
  \hline
  2 &3.06 &2.63 &1.08 &5.10 &3.61 &100\% &100\% \\
  \hline
  3 &0.81 &0.92 &0.34 &1.07 &1.12 &72\% &100\% \\
  \hline
  4 &1.00 &1.09 &0.40 &1.32 &1.26 &73\% &100\% \\
  \hline
  5 &0.18 &0.35 &0.09 &0.21 &0.28 &0\%   &100\% \\
  \hline
  All &1.00 & 1.08 &0.41 &1.39 &1.30 &74\% &100\% \\
  \hline
  3 - no failures &1.00 &1.11 &0.42 &1.32 &1.39 &100\% &100\% \\
  \hline
  4 - no failures  &1.22 &1.31 &0.49 &1.61 &1.53 &100\% &100\% \\
  \hline
  All - no failures &1.23&1.30 &0.5 &1.72 &1.60 &100\% &100\% \\
  \hline
\end{tabular}
\caption[Results]{Experimental results comparing $\mathcal{ATI}$ with $\mathcal{TP}$ in different scenarios. Values of all performance measures (except accuracy) for $\mathcal{TP}$ are expressed as a fraction of the values of the same measures for $\mathcal{ATI}$. $\mathcal{ATI}$ improves accuracy and computational efficiency, especially in dynamic domains.}
\label{tab:sim-results}
\end{table*}

Scenario-1 represents a standard planning task with no unexpected
domain changes. In this scenario, both $\mathcal{TP}$ and
$\mathcal{ATI}$ provide the same accuracy ($100\%$) and compute
essentially the same plan, but computing an activity comprising
intentional actions and repeatedly checking the validity of this
activity takes longer. This explains the reported average values of
$0.45$ and $0.81$ for planning time and total time (for
$\mathcal{TP}$) in Table~\ref{tab:sim-results} above.

In Scenario-2 (unexpected success), both $\mathcal{TP}$ and
$\mathcal{ATI}$ achieve $100\%$ accuracy. Here, $\mathcal{ATI}$ stops
reasoning and execution once it realizes the desired goal has been
achieved unexpectedly. However, $\mathcal{TP}$ does not realize this
because it does not consider observations not directly related to the
action being executed; it keeps trying to find the objects of interest
in different places. This explains why $\mathcal{TP}$ has a higher
planning time and execution time, computes more plans, and executes
more actions (i.e., plan steps) than $\mathcal{ATI}$.

Scenarios 3-5 correspond to different kinds of unexpected failures.
In each trial for these scenarios, $\mathcal{ATI}$ leads to a
successful achievement of the goal, whereas there are many instances
in which $\mathcal{TP}$ is unable to recover from the unexpected
observations and achieve the goal. For instance, if the goal is to
move two books to the library, and one of the books is moved to an
unexpected location when it is no longer part of an un-executed action
in the robot's plan, the robot may not reason about this unexpected
occurrence and the desired goal may not be achieved. This phenomenon
is especially pronounced in Scenario-5 that represents an extreme case
in which the robot using $\mathcal{TP}$ is never able to achieve the
assigned goal because it never realizes that it has failed to achieve
the goal. Notice that in the trials corresponding to all three
scenarios, $\mathcal{ATI}$ takes more time than $\mathcal{TP}$ to plan
and execute the plans for any given goal, but this increase in time is
justified given the high accuracy and the desired behavior that the
robot is able to achieve in these scenarios using $\mathcal{ATI}$.

The row labeled ``All'' in Table~\ref{tab:sim-results} shows the
average of the results obtained in the different scenarios. The
subsequent three rows in Table~\ref{tab:sim-results} summarize results
after removing from consideration trials in which $\mathcal{TP}$ fails
to achieve the assigned goal.  We then notice that $\mathcal{ATI}$ is
at least as fast as $\mathcal{TP}$ and is often faster, i.e., it takes
less time (overall) to plan and execute actions to achieve the desired
goal. In summary, $\mathcal{TP}$ may result in faster planning in
well-defined domains with little or no dynamic changes, but it results
in lower accuracy and higher execution time than $\mathcal{ATI}$ in
dynamic domains, especially in the presence of unexpected successes
and failures that are common in dynamic domains. The results in
Table~\ref{tab:sim-results} provide evidence in support of hypotheses
$H1$ and $H2$. The subsequent analysis of the fine-resolution
components of our architecture was thus performed by combining them
with $\mathcal{ATI}$ and not with $\mathcal{TP}$.

Next, to evaluate hypothesis $H3$, we ran experiments in the eight
complexity levels listed in Section~\ref{sec:expres}, with and without
including zooming. All trials included $\mathcal{ATI}$ for
coarse-resolution reasoning with the adapted theory of intentions, and
refined domain representation for fine-resolution reasoning. Recall
that the robot cannot actually execute the coarse-resolution actions.
As before, the goal in each experimental trial was to find and move a
target object to a target location. Similar to the experiments used to
evaluate $H1$ and $H2$, the values of performance measures without
zooming are, wherever appropriate, expressed a fraction of the values
with zooming. Table~\ref{tab:results-with-without-zoom} and
Table~\ref{tab:results-zoom-morecomplex} summarize the corresponding
results, and we make the following observations:
\begin{itemize}
\item When $\mathcal{ATI}$ was used with zooming, all trials in all
  complexity levels terminated successfully, i.e., the assigned goal
  was always achieved---see Table~\ref{tab:results-with-without-zoom}.
  Without zooming, the goal was achieved in all trials in complexity
  levels $L1$ and $L2$, in only $65\%$ of the trials in complexity
  level $L3$, and in none of the trials in complexity levels $L4-L8$.
  The observed failures in complexity levels $L3-L8$ were due to the
  existence of too many options (i.e., paths in the transition
  diagram) to consider during fine-resolution reasoning in the absence
  of zooming. In such cases, fine-resolution planning terminated
  unexpectedly (i.e., before the goal was achieved) in the absence of
  zooming. Thus, Tables~\ref{tab:results-with-without-zoom}
  and~\ref{tab:results-zoom-morecomplex} do not consider paired trials
  at or above complexity level $L4$; at these complexity levels, we
  only report results of trials that included zooming in
  fine-resolution reasoning.

\item The coarse-resolution reasoning time, i.e., the time for
  coarse-resolution planning and diagnostics, increases gradually (as
  expected) with the increase in the complexity level. In general, the
  time taken for coarse-resolution reasoning is much smaller in
  comparison with the fine-resolution reasoning time in complex
  domains The fine-resolution reasoning time, i.e. the time for
  planning at the fine resolution, and for inferring coarse-resolution
  observations based on fine-resolution outcomes, also increases with
  the increase in the complexity level. With zooming included in the
  fine-resolution reasoning, this increase is reasoning time scales
  well with the increase in the complexity level. However, in the
  absence of zooming, the increase in reasoning time is much more
  pronounced, e.g., fine-resolution reasoning at complexity level $L3$
  without zooming takes (on average) $55$ times as much time as when
  zooming is used.

\item Note that reasoning can imply multiple instances of planning and
  diagnostics for a particular goal. When zooming is used, the average
  time spent computing each refined plan scales well with the increase
  in the level of complexity. When zooming is not included in the
  fine-resolution reasoning, the average time spent in each refined
  plan increases dramatically, e.g., even at complexity level $L3$,
  each refined plan without zooming takes (on average) $85$ times as
  much time as with zooming.

\item The results with complexity levels $L7$ and $L8$ present an
  interesting comparison, and further indicate the benefits of
  zooming. Complexity level $L8$ has the same number of rooms and
  cells in each room as $L7$, but it has twice as many objects as
  $L7$. This increase would typically have caused a significant
  increase in the reasoning time, especially when we consider the
  parts of the different objects in the fine-resolution. However,
  zooming enables the robot to limit its attention to only the objects
  and object parts relevant to any given task; we only observe a small
  increase in the coarse-resolution reasoning time, with hardly any
  change in the fine-resolution reasoning time.
\end{itemize}
Overall, Tables~\ref{tab:results-with-without-zoom}
and~\ref{tab:results-zoom-morecomplex} indicate that zooming supports
scalable fine-resolution reasoning with the increase in complexity.
When used in conjunction with the $\mathcal{ATI}$ at the coarse
resolution, we obtain reliable and efficient performance in dynamic
domains. These results thus support hypothesis $H3$.

\begin{table}[tb]
\centering
\small\addtolength{\tabcolsep}{-3pt}
\begin{tabular}{|c|c|c|c|c|}
\hline
\multicolumn{2}{|c|}{\textbf{Complexity level}} & \textbf{L1} & \textbf{L2} & \textbf{L3} \\ \hline
\multirow{2}{*}{\textbf{Reasoning time (total)}} 
& Zoom     & 1.00 ($6.24\pm 1.56$) & 1.00 ($8.82\pm 2.95$) & 1.00 ($11.59\pm 3.88$)   \\ \cline{2-5} 
& No zoom & 1.01 ($6.28\pm 1.57$) & 1.20 ($10.74\pm 4.14$) & 20.07 ($225.49\pm 177.64$)      \\ \cline{1-5} %
\multirow{2}{*}{\textbf{Reasoning time (fine)}}
& Zoom     & 1.00 ($1.57\pm 0.51$) & 1.00 ($2.53\pm 0.92$) & 1.00 ($4.19\pm 1.48$)   \\ \cline{2-5} 
& No-zoom & 1.02 ($1.61\pm 0.54$) & 1.71 ($4.46\pm 2.29$)  & 55.23 ($218.1\pm 176.97$)    \\ \cline{1-5} %
\multirow{2}{*}{\textbf{Reasoning time (coarse)}}
& Zoom     & 1.00 ($4.67\pm 1.05$) & 1.00 ($6.29\pm 2.06$) & 1.00 ($7.4\pm 2.57$)  \\ \cline{2-5} 
& No zoom & 1.00 ($4.67\pm 1.03$) & 1.00 ($6.28\pm 2.06$) & 1.00 ($7.39\pm 2.58$)      \\ \cline{1-5} %
\multirow{2}{*}{\textbf{Time per refined plan}}
& Zoom     & 1.00 ($0.37\pm 0.02$) & 1.00 ($0.41\pm 0.02$)  & 1.00 ($0.56\pm 0.05$)  \\ \cline{2-5} 
& No zoom & 1.03 ($0.39\pm 0.01$) & 2.26 ($0.93\pm 0.32$)  & 83.49 ($45.98\pm 43.05$)       \\ \hline %
\multicolumn{1}{|c|}{\multirow{2}{*}{\textbf{Completed trials}}}
& Zoom     & $100\%$ & $100\%$ & $100\%$   \\ \cline{2-5} 
\multicolumn{1}{|c|}{} & No zoom & $100\%$ & $100\%$ & $65\%$ \\ \hline
\end{tabular}
\caption[A]{Performance with and without zooming at complexity levels $L1-L3$. Values of reasoning times without zooming are expressed as a fraction of the values with zooming. We only compute the ratio of reasoning times in trials that resulted in successful achievement of the assigned goal; this considers all the trials for complexity levels $L1-L2$, but only $65\%$ of the trials under $L3$. }
  \label{tab:results-with-without-zoom}
\end{table}

\begin{table}[tbh]
\centering
\small\addtolength{\tabcolsep}{-3pt}
\begin{tabular}{|c|l|l|l|l|l|}
\hline
 \textbf{Complexity level} & \textbf{L4} & \textbf{L5} & \textbf{L6} & \textbf{L7} & \textbf{L8} \\ \hline
\textbf{Reasoning time (total)}	& $18.63\pm 5.49$ & $22.73\pm 7.46$ & $29.8\pm 10.55$ & $41.93\pm 17.26$ 	& $42.99\pm 18.04$ \\   \hline
\textbf{Reasoning time (fine)}   & $8.89\pm 2.99$ & $10.78\pm 4.08$ & $17.77\pm 7.55$ & $30.07\pm 15.48$	& $29.36\pm 15.76$ \\   \hline
\textbf{Reasoning time (coarse)}	& $9.74\pm 3.26$ & $11.95\pm 4.17$ & $12.03\pm 4.33$ & $11.86\pm 4.12$ & $13.63\pm 5.07$ \\  \hline
\textbf{Time per refined plan} & $0.95\pm 0.26$ & $0.98\pm 0.22$ & $1.64\pm 0.68$ & $2.3\pm 1.02$ & $2.3\pm 1.09$ \\  \hline
\end{tabular}
\caption[B]{Reasoning time for trials using zooming at complexity levels $L4-L8$; trials without zooming were unable to terminate at these complexity levels. All trials using zooming were able to draw inferences and generate plans that achieved the assigned goal at all complexity levels. The increase in reasoning time with the increase in complexity levels is reasonable and demonstrates the scalability of our approach that combines $\mathcal{ATI}$ with refinement and zooming.}
  \label{tab:results-zoom-morecomplex}
\end{table}

\subsection{Experimental Results on Physical Robots}
\label{sec:expres-robots}
We also ran experimental trials with the combined architecture, i.e.,
$\mathcal{ATI}$ with refinement and zooming, on two different robot
platforms. These trials represented instances of the different
scenarios (in Section~\ref{sec:arch-toi}) in domains that are variants
of the RA domain in Example~\ref{ex:illus-example}.

First, consider the experiments with the Baxter robot manipulating
objects on a tabletop as shown in Figure~\ref{fig:example-robots}.
This domain is characterized by the following:
\begin{itemize}
\item The goal is to move particular objects between different
  ``zones'' (instead of places), or between particular cell locations
  within the zones, on a tabletop.
\item After refinement, each zone is magnified to obtain grid cells.
  Also, each object is magnified into parts such as $base$ and $top$
  after refinement.
\item Objects are characterized by the attributes $color$ and $size$.
\item The robot does not have a mobile base but it uses its arm to
  move objects between cells or zones.
\end{itemize}
Next, consider the experiments with the Turtlebot robot operating in
an indoor domain as shown in Figure~\ref{fig:example-robots}. This
domain is characterized by the following details:
\begin{itemize}
\item The goal is to find and move particular objects between places
  in an indoor domain.
\item The robot does not have a manipulator arm. It solicits help from
  a human to pickup the desired object when it has reached the
  location of the target object, and to put the object down when it
  has reached the location where it has to move the object.
\item Objects are characterized by the attributes $color$ and $type$.
\item After refinement, each place or zone was magnified to obtain
  grid cells. Also, each object is magnified into parts such as $base$
  and $handle$ after refinement.
\end{itemize}
Although the two domains differ significantly, e.g., in terms of the
domain attributes, actions and complexity, no change is required in
the architecture or the underlying methodology. Other than providing
the domain-specific information, no human supervision is necessary;
most of the other steps are automated. Similar to the experiments in
simulation, we used accuracy (of task completion) and time (for
planning and execution) as the performance measures, expressing the
values of relevant measures (e.g., planning time) for the baseline
implementation as a fraction of the values with our architecture. In
$\approx 50$ experimental trials in each domain, the robot using the
combined architecture is able to successfully achieve the assigned
goal. The performance is similar to that observed in the simulation
trials. For instance, if we do not include $\mathcal{ATI}$, i.e., use
$\mathcal{TP}$ with refinement and zooming, the accuracy with which
the goal is achieved reduces from $100\%$ to $\approx 60\%$, and it
takes $\approx 2-3$ times as much time to achieve the goal, especially
in the presence of unexpected success or failure. In other scenarios,
the performance with $\mathcal{ATI}$ is at least as good as that with
$\mathcal{TP}$.  Also, if we do not include zooming, the robot takes
significantly longer to plan and execute concrete actions.  In fact,
as the domain becomes more complex, i.e., with an increase in the
number of domain objects and the length of the plan required to
achieve the desired goal, planning starts becoming computationally
expensive and (often) computationally unfeasible without zooming.
These results support the three hypotheses listed in
Section~\ref{sec:expres}.

Videos of the trials on the Baxter robot and Turtlebot corresponding
to different scenarios can be viewed online~\cite{video-results}.
For instance, in one trial involving the Turtlebot, the goal is to
have both a cup and a bottle in the $library$, and these objects and
the robot are initially in $of\!\!fice_2$. The computed plan has the
robot pick up the bottle, move to the $kitchen$, move to the
$library$, put the bottle down, move back to the $kitchen$ and then to
$of\!\!fice_2$, pick up the cup, move to the $library$ through the
$kitchen$, and put the cup down. When the Turtlebot is moving to the
$library$ holding the bottle, someone moves the cup to the $library$.
With $\mathcal{ATI}$, the robot uses the observation of the cup (as it
is putting the bottle down in the $library$), to infer that the goal
has been achieved and to terminate plan execution early. Without
$\mathcal{ATI}$, i.e., with $\mathcal{TP}$, the robot continued with
its initial plan and realized that there was a problem (unexpected
observation of the cup in the $library$) only when it went back to
$of\!\!fice_2$ and did not find the cup there.

Similarly, in one trial with the Baxter, the goal is to have blue
blocks and green blocks in zone Y (\emph{yellow} zone on the right
side of the screen) and these blocks are initially in zone R
(\emph{red} zone on the left side of the screen). The computed plan
has the Baxter move its arm to zone R, pick up a block, move to zone G
(\emph{green} zone in the center) then to zone Y to put the block
down, and repeat this process until it has moved all the blue blocks
and green blocks. When the Baxter has moved one block and is moving
back to pick up the second block from zone R, an exogenous action puts
the first block in zone G. With $\mathcal{ATI}$, as the Baxter is
moving over zone G on the way to zone R, it observes the block (it had
previously put in zone Y), performs diagnostics and realizes his
current activity will not achieve the goal. It then stops executing
its current activity, computes a new activity of intentional actions,
and succeeds in moving both blocks to zone Y. With $\mathcal{TP}$, the
robot is not able to use the observation of the first block in zone G,
continues with the initial plan and never realizes that the goal has
not been achieved.

\section{Discussion and Future Work}
\label{sec:conclusions}
In this paper we presented a general architecture that represents and
reasons with intentional actions. The architecture represents and
reasons with domain knowledge and beliefs encoded as tightly-coupled
transition diagrams at two different resolutions, with the
fine-resolution description defined as a refinement of the
coarse-resolution description. For any given goal, non-monotonic
logical reasoning with the coarse-resolution domain representation
containing commonsense domain knowledge is used to provide a plan of
intentional abstract actions. The coarse-resolution transition
corresponding to each such abstract intentional action is implemented
as a sequence of concrete actions by automatically identifying and
reasoning with the part of the fine-resolution representation relevant
to the coarse-resolution transition and the coarse-resolution goal.
The execution of each concrete action uses probabilistic models of the
uncertainty in sensing and actuation, and any associated outcomes are
added to the coarse-resolution history. Experimental results in
simulation and on different robot platforms, as summarized above,
indicate that this architecture improves the accuracy and
computational efficiency of decision making in comparison with an
architecture that does not reason with intentional actions. It also
significantly improves the computational efficiency of decision making
in comparison with an architecture that does not support zooming in
the fine resolution.

This architecture opens up multiple directions for future research
that build on the capabilities of the current architecture. First,
although the current architecture builds on key results of the
coupling between the transition diagrams, it will be interesting to
formally establish the relationship between the different transition
diagrams in this architecture, along the lines of the analysis
provided in~\cite{mohan:JAIR19}. This will enable any designer using
our architecture for a particular robotics domain to establish
correctness of the algorithms and build trust in the resultant
behavior of the robot. Second, the results reported in this paper are
based on experimental trials in variants of a particular (RA) domain.
However, the underlying capability of modeling and reasoning about
intentional actions is relevant to other problems and applications
characterized by dynamic changes. For instance, other work within our
research group has combined the reasoning capabilities of our
architecture with inductive learning of domain constraints to guide
the construction of deep networks that have been used for estimating
the occlusion and stability of object structures~\cite{mota:rss19} and
for answering explanatory questions about
images~\cite{riley:hai18,riley:rainr19}; other research groups have
explored the combination of ASP-based knowledge representation with
low-level perceptual processing for explaining spatial relations in
videos~\cite{suchan:aaai18}. Future research can adapt our
architecture to such problems in more complex domains to demonstrate
the scalability and wider applicability of our architecture. Third,
the relational representation and reasoning capabilities supported by
our architecture can be used to provide explanations of the decisions
made, the underlying beliefs, and the experiences that informed these
beliefs. Currently, our architecture only reasons with representations
at two different resolutions, but proof of concept work indicates that
it is possible to introduce a theory of explanations and expand the
notion of refinement to interactively provide explanations at
different levels of abstraction~\cite{mohan:aaaisymp19}. Fourth, the
architecture has only considered a single robot representing and
reasoning with intentional actions. There is considerable research on
a team of robots working with humans, including approaches based on
logic programming approaches~\cite{erdem:KI18}. Future work can extend
our architecture to a team of robots collaborating with humans in
dynamic application domains such as disaster rescue, surveillance, and
healthcare.  The long-term goal is to enable a team of robots
collaborating with humans in complex domains to represent and reason
reliably and efficiently with different descriptions of incomplete
domain knowledge and uncertainty.


\section*{Acknowledgements}
  The authors thank Michael Gelfond for discussions related to the
  modeling of defaults and exogenous actions in the architecture
  reported in this paper.


\bibliographystyle{plain}
\bibliography{bibliography}


\end{document}